\documentclass[10pt,twocolumn,letterpaper]{article}

\usepackage{cvpr}
\usepackage{times}
\usepackage{epsfig}
\usepackage{graphicx}
\usepackage{amsmath}
\usepackage{amssymb}
\usepackage[misc]{ifsym}
\usepackage{appendix}
\usepackage{subfigure}
\usepackage{booktabs}

\usepackage{hyperref}

\cvprfinalcopy 

\def\cvprPaperID{****} 

\begin{document}


\title{3D Anisotropic Hybrid Network: Transferring Convolutional Features from 2D Images to 3D Anisotropic Volumes}

\author{Siqi Liu\textsuperscript{1*},
Daguang Xu\textsuperscript{2*},
S. Kevin Zhou\textsuperscript{2\Letter},
Thomas Mertelmeier\textsuperscript{3},
Julia Wicklein\textsuperscript{3},\\
Anna Jerebko\textsuperscript{3},
Sasa Grbic\textsuperscript{2}, 
Olivier Pauly\textsuperscript{2},
Weidong Cai\textsuperscript{1}, Dorin Comaniciu\textsuperscript{2}\\
\textsuperscript{1}School of Information Technologies, University of Sydney, Australia\\
\textsuperscript{2}Siemens Healthineers, Medical Imaging Technologies, Princeton, NJ, USA\\
\textsuperscript{3}Siemens Healthineers, X-ray products, Germany\\
\textsuperscript{*}Authors contributed equally\\
\textsuperscript{\Letter}Corresponding author
\href{mailto:shaohua.zhou@siemens-healthineers.com}{shaohua.zhou@siemens-healthineers.com}
}

\maketitle

\begin{abstract}
While deep convolutional neural networks (CNN) have been successfully applied for 2D image analysis, it is still challenging to apply them to 3D anisotropic volumes, especially when the within-slice resolution is much higher than the between-slice resolution and when the amount of 3D volumes is relatively small. On one hand, direct learning of CNN with 3D convolution kernels suffers from the lack of data and likely ends up with poor generalization; insufficient GPU memory limits the model size or representational power. On the other hand, applying 2D CNN with generalizable features to 2D slices ignores between-slice information. Coupling 2D network with LSTM to further handle the between-slice information is not optimal due to the difficulty in LSTM learning. To overcome the above challenges, we propose a 3D Anisotropic Hybrid Network (AH-Net) that transfers convolutional features learned from 2D images to 3D anisotropic volumes. Such a transfer inherits the desired strong generalization capability for within-slice information while naturally exploiting between-slice information for more effective modelling. The focal loss is further utilized for more effective end-to-end learning. We experiment with the proposed 3D AH-Net on two different medical image analysis tasks, namely lesion detection from a Digital Breast Tomosynthesis volume, and liver and liver tumor segmentation from a Computed Tomography volume and obtain the state-of-the-art results.
\end{abstract}

\begin{figure}
  \centering
  \includegraphics[width=1\linewidth]{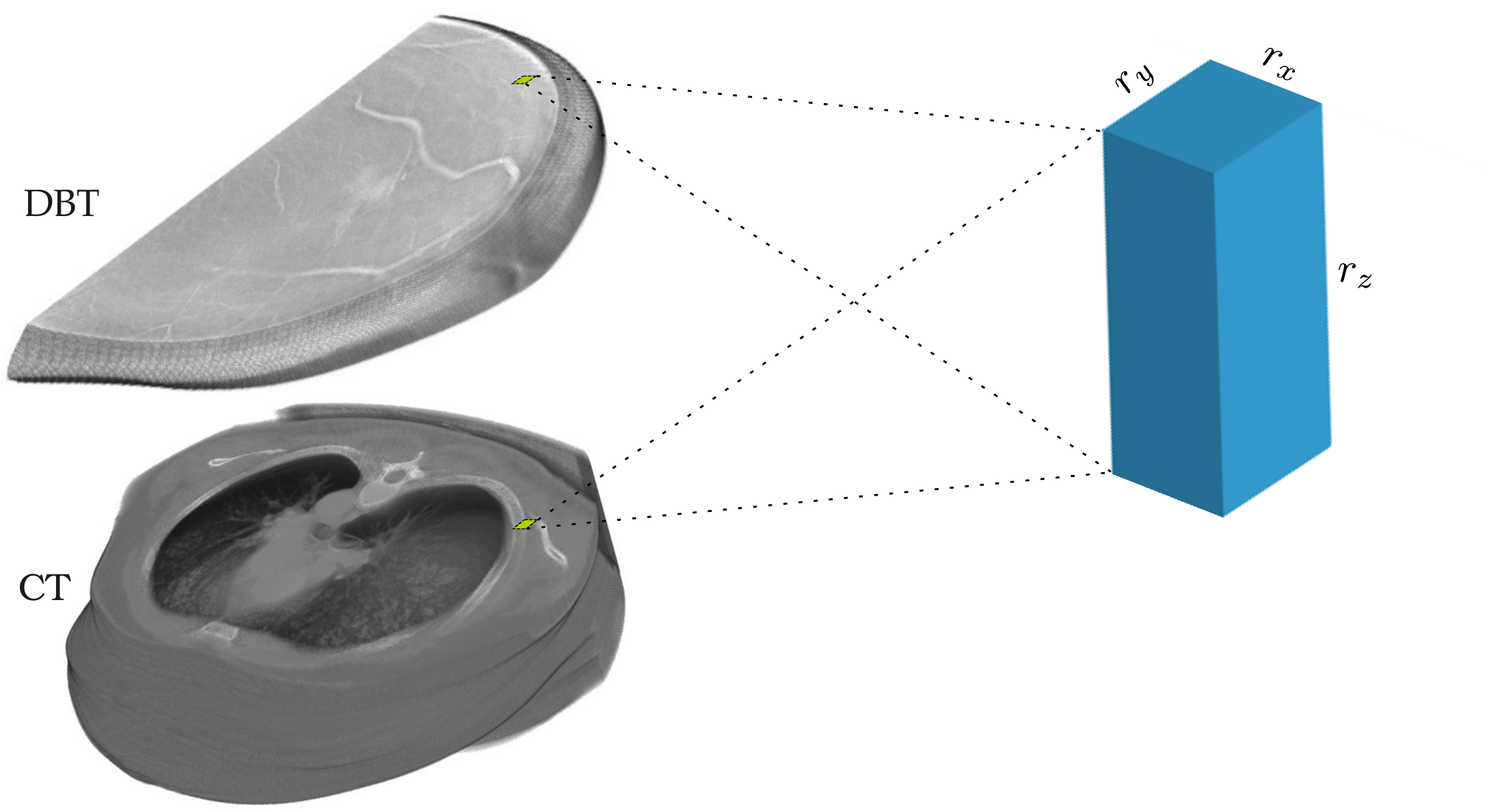}
  \caption{The example anisotropic volumes of DBT and CT are shown in the left column. Such volumes contain voxels with much higher within-slice resolution $r_x\times r_y$ than the between-slice resolution $r_z$.}
  \label{fig:gcn}
\end{figure}

\section{Introduction}
3D volumetric images (or volumes) are widely used for clinical diagnosis, surgical planning and biomedical research. The 3D context information provided by such volumetric images are important for visualising and analysing the object of interest. However, given the added dimension, it is more time consuming and sometimes harder to interpret 3D volumes than 2D images by machines. Many previous studies use convolutional neural networks (CNN) to extract the representation of structural patterns of interests in human or animal body tissues.

Due to the special imaging settings, many imaging modalities come with anisotropic voxels, meaning not all the three dimensions have equal resolutions.
For examples, in the 3D volumes of Digital Breast Tomosynthesis (DBT), and sometimes Computed Tomography (CT), the image resolution in $xy$ plane/slice (or within-slice resolution) is more than ten times higher than that of the $z$ resolution (or between-slice resolution).
Thus, the $xy$ slices preserve much more information than the $z$ dimension.
In DBT images, only the spatial information within the $xy$ plane can be guaranteed. However, the 3D context between $xy$ slices, even with slight misalignment, still carries meaningful information for analysis.

Directly applying 3D CNN to such images remains a challenging task due to the following reasons:
(1) It may be hard for a small $3\times3\times3$ kernel to learn useful features from anisotropic voxels, because of the different information density along each dimension.
(2) Theoretically more features are needed in 3D networks compared to 2D networks. The capability of 3D networks is bounded by the GPU memory, constraining both the width and depth of the networks.
(3) Unlike 2D computer vision tasks which nowadays can make use of the backbone networks pretrained using millions of 2D images \cite{Russakovsky2015ImageNetChallenge}, 3D tasks mostly have to train from scratch , and hence suffer from the lack of large 3D datasets. In addition, the high data variations make the 3D networks harder to be trained.
Also, 3D CNNs trained on such small image datasets with relatively small 3D context are hard to generalize to unseen data.

Besides the traditional 3D networks built with $1\times1\times1$ and $3\times3\times3$ kernels, there are other methods for learning representations from anisotropic voxels.
Some studies process 2D slices separately with 2D networks \cite{Liu2017DeepImaging}.
To make a better use of the 3D context, more than one image slice is used as the input for 2D networks \cite{Li2017H-DenseUNet:Volumes,Zeng2017DeepEM3D:Segmentation}.
The 2D slices can also be viewed sequentially by combining a fully convolutional network (FCN) architecture with Convolutional LSTM to view the adjacent image slices as a time series to distil the 3D context from a sequence of abstracted 2D context \cite{Chen2016CombiningSegmentation}.
There are also a few studies using anisotropic convolutional kernels to distribute more learning capability on the $xy$ plane than on the $z$ axis \cite{Brosch2016DeepSegmentation,Lee2017SuperhumanChallenge,Wang2017AutomaticNetworks}.

In this paper, we propose the 3D Anisotropic Hybrid Network (AH-Net) to learn informative features from images with anisotropic resolution.
To obtain the 3D AH-Net, we firstly train a 2D fully convolutional ResNet \cite{Peng2017LargeNetwork} which is initialized with pre-trained weights and uses multiple 2D image slices as inputs.
The feature encoder of such a 2D network is then transformed into a 3D network by extending the 2D kernel with one added dimension.
Then we add a feature decoder sub-network to extract the 3D context.
The feature decoder consists of anisotropic convolutional blocks with $3\times3\times1$ and $1\times1\times3$ convolutions.
Different anisotropic convolutional blocks are combined with dense connections \cite{Huang2016DenselyNetworks}.
Similar to U-Net \cite{Ronneberger2015U-Net:Segmentation}, we use skip connections between the feature encoder and the decoder.
A pyramid volumetric pooling module \cite{Zhao2016PyramidNetwork} is stacked at the end of the network before the final output layer for extracting multiscale features.

Since AH-Net can make use of 2D networks pre-trained with large 2D general image datasets such as ImageNet \cite{Russakovsky2015ImageNetChallenge}, it is easier to train as well as to generalize.
The anisotropic convolutional blocks enable it to exploit the 3D context.
With end-to-end inference as a 3D network, AH-Net runs much faster than the conventional multi-channel 2D networks regarding the GPU time required for processing each 3D volume.

\section{Related Work}
It is hard for conventional 3D neural networks with isotropic $3\times 3\times 3$ kernels to extract robust representations from 3D volumes with anisotropic resolution.
The most intuitive approach is to re-sample the images to isotropic resolutions \cite{Moeskops2016AutomaticNetwork}. 
This would work when the difference between three dimensions are small, and the spatial information between different slices is accurate. When the $z$ resolution is much smaller than the $xy$ resolution, the majority of voxels added by image resampling are redundant, thus introducing unnecessary extra computational cost. It may also result in loss of information if downsampling happens in the $xy$ direction.

Instead of using 3D networks, some studies deal with the voxel anisotropy using 2D networks. DeepEM3D-Net \cite{Zeng2017DeepEM3D:Segmentation} has only two 3D convolution layers to integrate 3D information in the early stages and performs 2D convolution for the rest of the following layers in an FCN. The input to DeepEM3D-Net is a stack of 2D image slices. The resultant 3D segmentation is obtained by concatenating the 2D output slices. HDenseNet \cite{Li2017H-DenseUNet:Volumes} applies 2D networks on all image slices at first. Then a 3D DenseUNet is applied on the concatenated 3D output volume to obtain the final result. Different from our proposed network, HDenseNet does not have shared convolutions between the 2D and 3D networks. Also, we use anisotropic 3D convolutional blocks to replace the isotropic 3D convolutions.

A bi-directional convolutional LSTM (BDC-LSTM) and an FCN model are combined to view slices as a time series \cite{Chen2016CombiningSegmentation}.
BDC-LSTM is trained to exploit the 3D contexts by applying a series of 2D convolutions on the $xy$ plane in a recurrent fashion to interpret 3D contexts while propagating contextual information in the $z$-direction. The FCN model is used for extracting the initial 2D feature maps which are used as the inputs to BDC-LSTM. The final output is obtained from the BDC-LSTM model with a softmax layer. Though the idea of fusing the 2D features to maintain the between-slice consistency is similar to our proposed method, we believe this can be achieved with stacked anisotropic convolution blocks, which are easier to train and to generalize than the convolutional LSTM.

Some studies use 3D convolutional kernels with anisotropic sizes to distribute more learning capability to the $xy$ plane. For example, $9\times9\times5$ convolutions are used in \cite{Brosch2016DeepSegmentation}. However, large convolution kernels would bring higher computational cost. 
Two more recent studies \cite{Qiu2017LearningNetworks,Wang2017AutomaticNetworks,Lee2017SuperhumanChallenge} use small kernels to simulate the large anisotropic kernels.
The convolution modules in \cite{Lee2017SuperhumanChallenge} starts with a $3\times1\times1$ convolution, followed by two $3\times3\times3$ convolutions.
Similar to our work, all the isotropic convolutions are replaced by $3\times3\times1$ and $1\times1\times3$ convolutions in \cite{Qiu2017LearningNetworks,Wang2017AutomaticNetworks}.
Several possible designs of combining the $3\times3\times1$ and $1\times1\times3$ kernels are discussed in a recent paper \cite{Wang2017AutomaticNetworks} that focuses on video learning.
Our network is different to the ones in \cite{Qiu2017LearningNetworks,Wang2017AutomaticNetworks} since we use the anisotropic 3D convolutions only in the feature decoder while the encoder is locked with pre-trained weights transferred from a 2D network.
It allows the proposed AH-Net to use any 2D fully convolutional networks pre-trained on large-scale datasets for initializing the encoder network.

\begin{figure}
  \centering
  \includegraphics[width=1\linewidth]{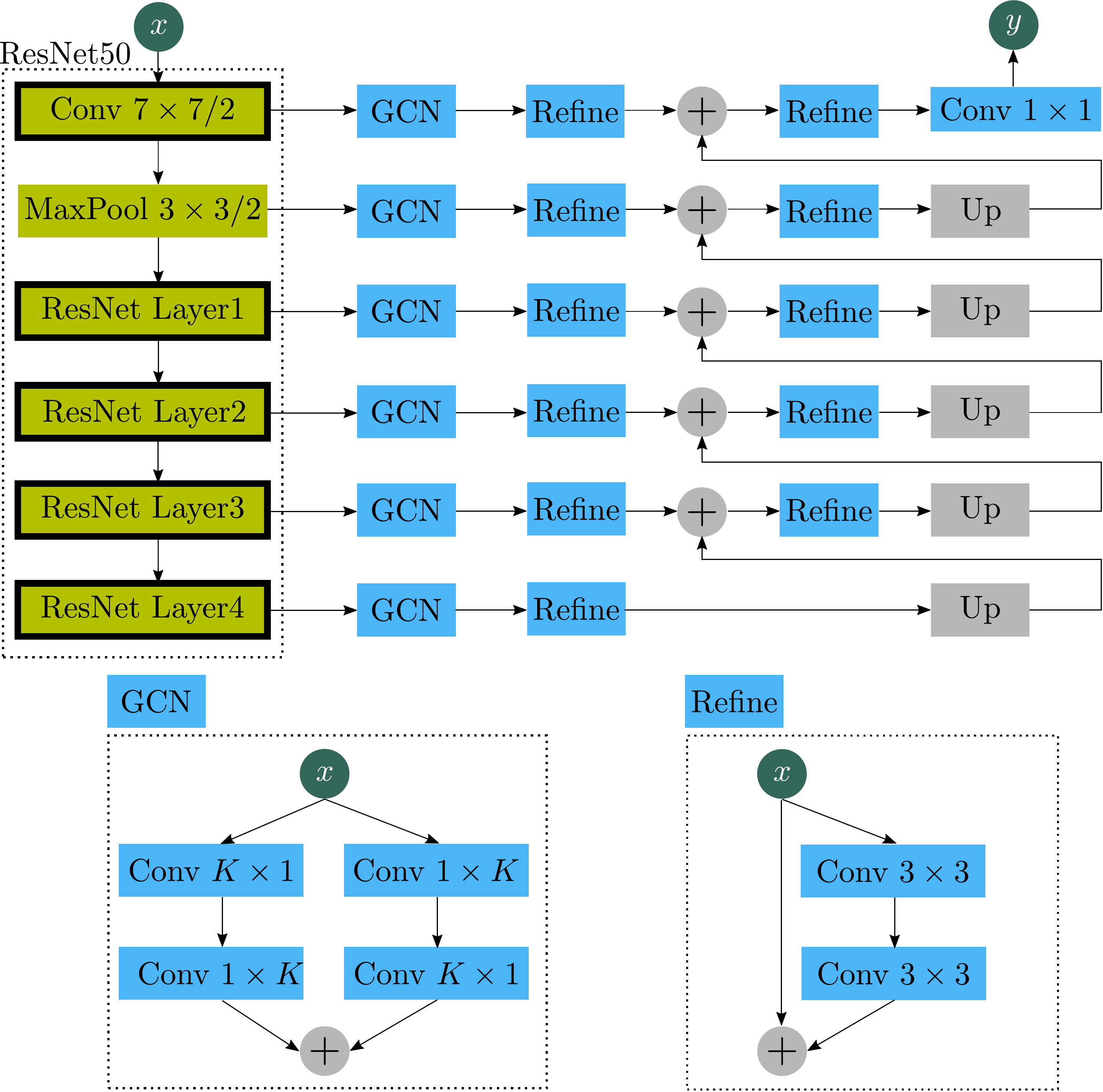}
  \caption{The network architecture for pre-training the 2D encoder network Multi-Channel Global Convolutional Neural Network (MC-GCN). The ResNet50 is used as the back-bone network, initialized with ImageNet images. The global convolutional network modules and refinement modules \cite{Peng2017LargeNetwork} are added to the encoder network to increase the receptive field during the pre-training as well as to increase the output response map to the original resolution. Conv $K\times K / S$ represents a convolution layer with the kernel size $K$ and the stride size $S$ in each dimension. The upsampling module (Up) consists of a Conv $1\times1$ projection layer and a bi-linear upsampling layer.}
  \label{fig:gcn}
\end{figure}

\begin{figure*}
  \centering
  \includegraphics[width=1\linewidth]{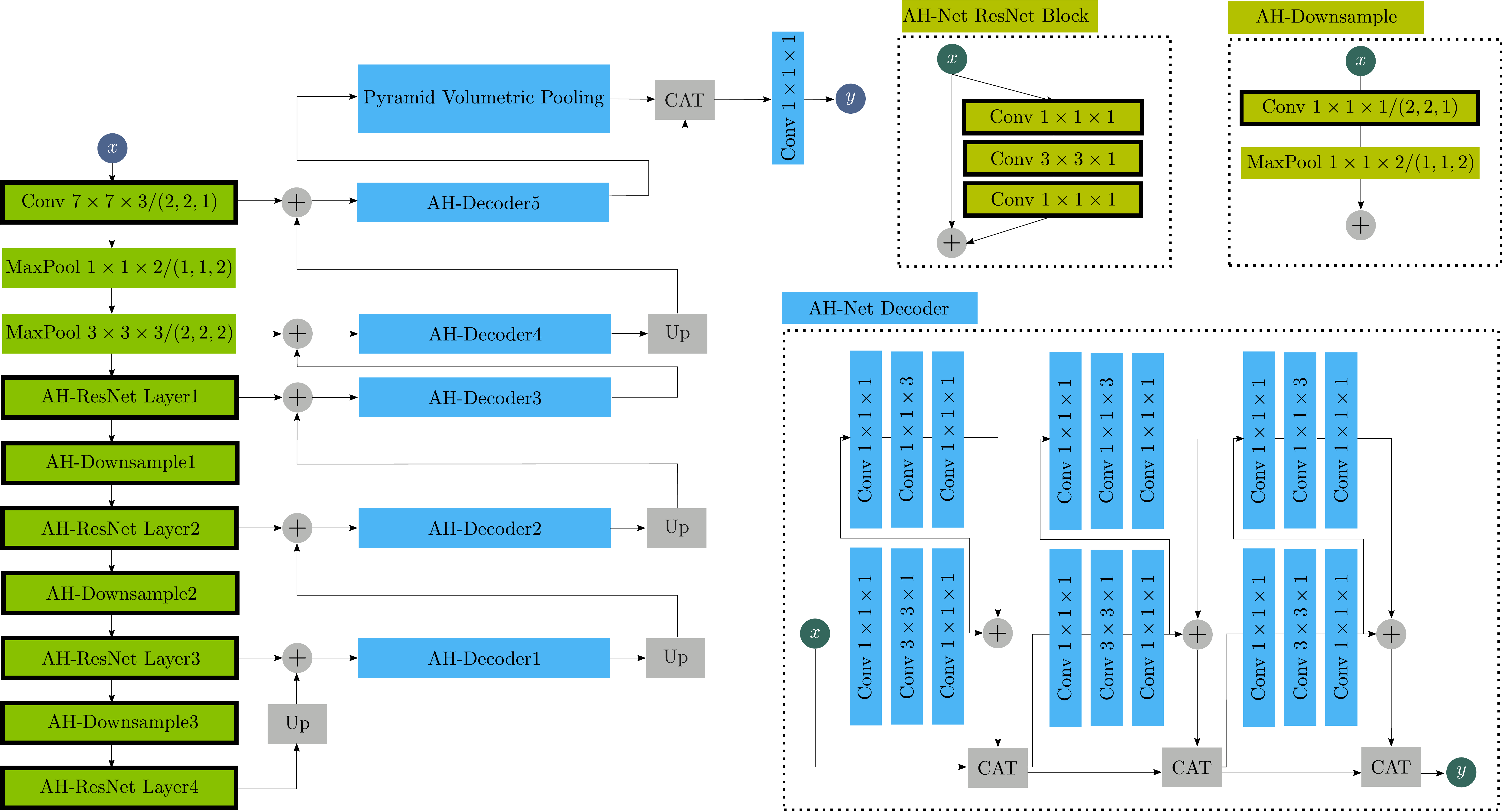}
  \caption{The architecture of 3D AH-Net. The feature encoder with AH-ResNet blocks is transferred from the pre-trained 2D network with $1\times1\times1$ and $3\times3\times1$ convolutions. The features are then processed with the AH-Net decoders which are designed with $3\times3\times1$ and $1\times1\times3$ convolutional blocks. Feature summation is used instead of concatenation as in \cite{Chaurasia2017LinkNet:Segmentation} to support more feature maps with less memory consumption. The pyramid pooling \cite{Zhao2016PyramidNetwork} is used for extracting the multiscale feature responses. We hide the batch normalization \cite{Ioffe2015BatchShift} and ReLu layers for brevity. The weights of the blocks with black borders are transformed from the 2D MC-GCN.}
  \label{fig:ahnet}
\end{figure*}

\begin{figure}
  \centering
  \includegraphics[width=1\linewidth]{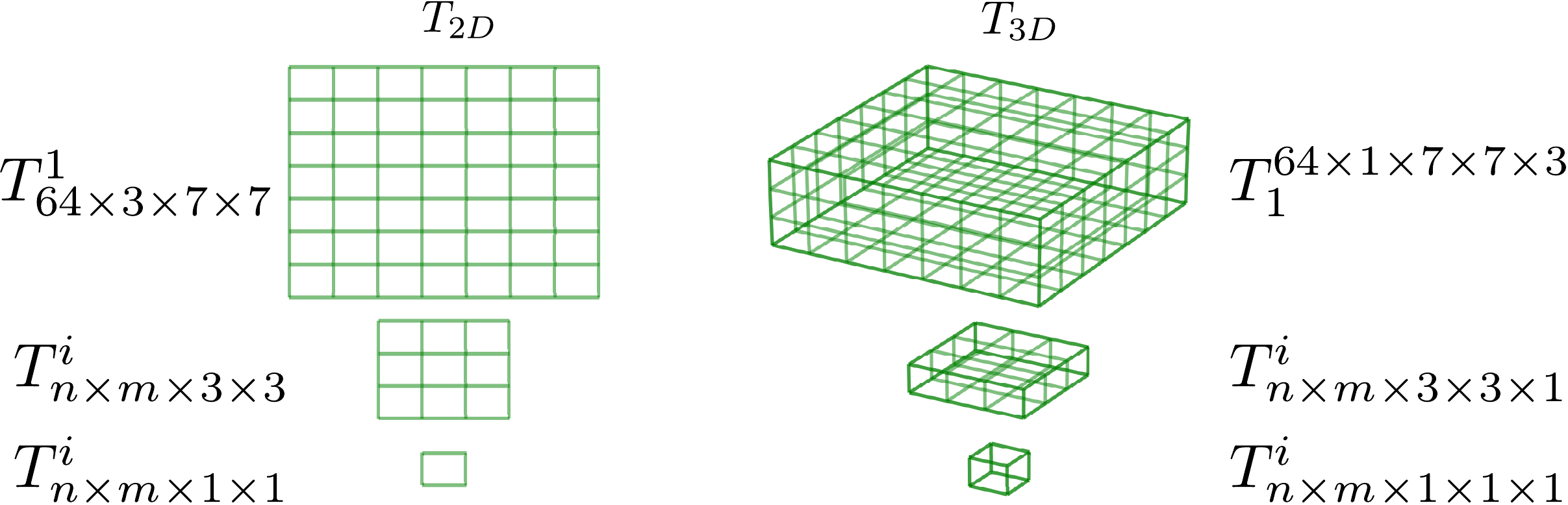}
  \caption{Transforming the 2D convolutional weight tensor $T^{2D}$ to 3D $T^{3D}$, where $m$ and $n$ are the number of features and channels of a layer, respectively. 
  The 1st layer weight tensor $T^1_{64 \times 3 \times 7 \times 7}$ is transformed to $T^1_{64\times 1 \times 7 \times 7 \times 3}$.
  The other convolutional kernels are transformed by adding an extra dimension.
  }
  \label{fig:kernels}
\end{figure}


\section{Anisotropic Hybrid Network}
The AH-Net consists of a feature encoder and a feature decoder. 
The encoder, transformed from a 2D network, is designed for extracting the deep representations from 2D slices with high resolution. 
The decoder built with densely connected blocks of anisotropic convolutions is responsible for exploiting the 3D context and maintaining the between-slice consistency. 
The network training is performed in two stages: 
the encoder is learned; 
then the 3D decoder is added and fine-tuned with the encoder parameters locked. 
To perform end-to-end hard voxel mining, we use the Focal Loss (FL) originally designed for object detection \cite{Lin2017FocalDetection}.

\subsection{Learning a multi-channel 2D feature encoder}
We train a 2D multi-channel global convolutional network (MC-GCN) similar to the architecture proposed in \cite{Peng2017LargeNetwork} to extract the 2D within-slice features at different resolutions, as shown in Fig.~\ref{fig:gcn}.
In this paper, we choose the ResNet50 model \cite{He2016} as the back-bone network which is initialized by pre-training with the ImageNet images \cite{Russakovsky2015ImageNetChallenge}, although other pre-trained networks would work similarly.
The network is then fine-tuned with 2D image slices extracted from the 3D volumes.
The input to this network is three neighbouring slices (treated as RGB channels). Thus, the entire architecture of the ResNet50 remains unchanged.
The multi-channel 2D input could enable the 2D network to fuse the between-slice context at an early stage.
A decoder is added to accompany the encoder to upscale the response map to the original resolution.
We choose the decoder architecture with the global convolutional networks (GCN) and refinement blocks \cite{Peng2017LargeNetwork}.
The GCN module simulates a large $K\times K$ convolutional kernel by decomposing it into two 1-D kernels ($1 \times K$ and $K \times 1$). Two branches containing the 1D kernels permuted in different orders are merged by summation.
The output of each GCN module contains the same number of output maps as the final outputs.
The large kernels simulated by GCNs ensure that the network has a large receptive field at each feature resolution.
Each refinement block contains two $3\times3$ convolutions with a ReLU activation in the middle.
The input of each refinement block is also added to its output to form a residual connection.
At the end of each encoder resolution level, the features are fed into GCN modules with the kernel sizes of $63, 31, 15, 9, 7,5$, respectively. The output features are fed into a refinement block and summed with the features upsampled from a lower resolution level.
The summed features are fed into another refinement block and upsampled with a $1\times1$ convolution and a bi-linear upsampling layer.
The final output has the same resolution as the image input.
The decoder has only a small number of parameters with little computational cost.
The light-weight decoder makes the encoder features easier to be transferred to the 3D AH-Net since the majority of the feature learning relies on the encoder network.

\subsection{Transferring the learned 2D net to 3D AH-Net}
\label{sec:AH-Net}
The architecture of the proposed 3D anisotropic hybrid network (AH-Net) is shown in Fig. \ref{fig:ahnet}. After the 2D MC-GCN network converges, we extract the parameters of its encoder and transfer them to the corresponding encoder layers of AH-Net. The decoder part of the 2D MC-GCN is discarded and instead we design a new decoder for the AH-Net that consists of multiple levels of densely connected blocks, followed by a pyramid volumetric pooling module. The parameters of the new decoder are randomly initialized. The input and output of AH-Net are now 3D patches, similar to other conventional 3D CNN.
The transformation of convolution tensors from 2D to 3D is illustrated in Fig.~\ref{fig:kernels}, which aims to perform 2D convolutions on 3D volumes slice by slice in the encoder part of AH-Net. 

\subsubsection{Notations}
A 2D convolutional tensor is denoted by $T^{i}_{n\times m\times h\times w}$, where $n$, $m$, $h$, and $w$ respectively represent the number of output channels, the number of input channels, the height and width of the $i^{th}$ convolution layer.
Similarly, a 3D weight tensor is denoted by $T^{i}_{n\times m\times h\times w \times d}$ where $d$ is the filter depth. 
We use $P^{(b,a,c,d)}(T_{a\times b \times c \times d})$ to denote the dimension permutation of a tensor $T_{a\times b \times c \times d}$, resulting in a new tensor $T_{b\times a \times c \times d}$ with the $1^{st}$ and $2^{nd}$ dimensions switched.
$P^{(a,\ast,b,c,d)}(T_{a\times b \times c \times d})$ adds an identity dimension between the $1^{st}$ and $2^{nd}$ dimensions of the tensor $T_{a\times b \times c \times d}$ and gives $T_{a\times 1 \times b \times c \times d}$. 
We define a convolutional layer as Conv $K_x \times K_y \times K_z / (S_x, S_y, S_z)$, where $K_x$, $K_y$ and $K_z$ are the kernel sizes; $S_x$, $S_y$ and $S_z$ are the stride step size in each direction.
Max pooling layers are denoted by MaxPool $K_x \times K_y \times K_z / (S_x, S_y, S_z)$.
The stride is omitted when a layer has a stride size of 1 in all dimensions.

\subsubsection{Input layer transform}
The input layer of the 2D MC-GCN contains a convolutional weight tensor $T^{1}_{64\times 3\times7\times7}$ inherited from its ResNet50 back-bone network.
The 2D convolutional tensor $T^{1}_{64\times 3\times7\times7}$ is transformed into 3D as 
\begin{equation}
P^{(1,*,3,4,2)}(T^{1}_{64\times3\times7\times7})=T^{1}_{64\times1\times7\times7\times3}
\end{equation}
in order to form a 3D convolution kernel that convolves $3$ neighbouring slices.
To keep the output consistent with the 2D network, we only apply stride-$2$ convolutions on the $xy$ plane and stride $1$ on the third dimension. 
This results in the input layer Conv $7\times 7 \times 3 / (2,2,1)$.
To downsample the $z$ dimension, we use a MaxPool $1\times1\times2 / (1,1,2)$ to fuse every pair of the neighbouring slices.
An additional MaxPool $3\times3\times3 / (2,2,2)$ is used to keep the feature resolution consistent with the 2D network.

\subsubsection{ResNet block transform}
All the 2D convolutional tensors $T^i_{n\times m\times1\times1}$ and $T^i_{n\times m\times3\times3}$ in the ResNet50 encoder are transformed as 
\begin{equation}
P^{(1,2,3,4,\ast)}(T^i_{n\times m\times1\times1})=T^i_{n\times m\times1\times1\times1}
\end{equation}
and 
\begin{equation}
    P^{(1,2,3,4,\ast)}(T^i_{n\times m\times3\times3})=T^i_{n\times m\times3\times3\times1}.
\end{equation}    
In this way, all the ResNet Conv $3\times 3 \times 1$ blocks as shown in Fig.~\ref{fig:ahnet} only perform 2D slice-wise convolutions on the 3D volume within the $xy$ plane.
The original downsampling between ResNet blocks is performed with Conv $1 \times 1 / (2, 2)$. 
However, in a 3D volume, a Conv $1\times 1 \times 1 / (2,2,2)$ skips a slice for every step on the $z$ dimension. This would miss important information when the image only has a small number of slices along the $z$ dimension, especially for detection tasks.
We therefore use a Conv $1\times 1\times 1 / (2,2,1)$ following by a MaxPool $1\times1\times2 / (1,1,2)$ to downsample the 3D feature maps between the ResNet blocks as shown in the AH-Downsample block in Fig.~\ref{fig:ahnet}.
This MaxPooling simply takes the maximum response along the $z$ direction between 2 neighbouring slices. 
Unlike the previous studies that avoided downsampling along the $z$ direction \cite{Lee2017SuperhumanChallenge}, we find it important for allowing the use of large and deep  networks on 3D data with limited GPU memory.

\subsection{Anisotropic hybrid decoder}
Accompanying to the transformed encoder, an anisotropic 3D decoder sub-network is added to exploit the 3D anisotropic image context.
In the decoder, anisotropic convolutional blocks with Conv $1\times1\times1$, Conv $3\times3\times 1$ and Conv $1\times 1\times 3$ are used.
The features are passed into an $xy$ bottleneck block at first with a Conv $3 \times 3 \times 1$ surrounded by two layers of Conv $1\times1\times1$.
The output is then forwarded to another bottleneck block with a Conv $1\times1\times3$ in the middle and summed with itself before forwarding to the next block.
This anisotropic convolution block decomposes a 3D convolution into 2D and 1D convolutions.
It receives the inputs from the previous layers using a 2D convolution at first, preserving the detailed 2D features.
Conv $1\times1\times3$ mainly fuses the within-slice features to keep the $z$ dimension output consistent.

Three anisotropic convolutional blocks are connected as the densely connected neural network \cite{Huang2016DenselyNetworks} using feature concatenation for each resolution of encoded features.
Similar to LinkNet \cite{Chaurasia2017LinkNet:Segmentation}, the features received from each resolution of the encoder are firstly projected to match the number of features of the higher encoder feature resolution using a Conv $1\times1\times 1$.
They are then upsampled using the 3D tri-linear interpolation and summed with the encoder features from a higher resolution.
The summed features are forwarded to the decoder blocks in the next resolution.

At the end of the decoder network, we add a pyramid volumetric pooling module \cite{Zhao2016PyramidNetwork} to obtain multi-scaled features. 
The output features of the last decoder block are firstly down-sampled using 4 different Maxpooling layers, namely MaxPool $64\times64\times1$, MaxPool $32\times32\times1$, MaxPool $16\times16\times1$ and MaxPool $8\times8\times1$ to obtain a feature map pyramid.
Conv $1\times1\times1$ layers are used to project each resolution in the feature pyramid to a single response channel.
The response channels are then interpolated to the original size and concatenated with the features before downsampling.
The final outputs are obtained by applying a Conv $1\times1\times1$ projection layer on the concatenated features.

\subsection{Training AH-Net using Focal Loss}
\label{sec:train}
Training AH-Net using the same learning rate on both the pre-trained encoder and the randomly initialized decoder would make the network difficult to optimize.
To train the 3D AH-Net, all the transferred parameters are locked at first.
Only the decoder parameters are fine-tuned in the optimization.
All the parameters can be then fine-tuned altogether afterwards to the entire AH-Net jointly.

The training of 3D fully convolution networks tend to pre-mature on the easy voxels quickly and converge slowly on the hard voxels, which are sometimes the objects of interests in medical images. 
For example, FCNs would learn the background voxels with uniform distributions quickly. 
For small-scaled patterns, such as lesions and object boundaries, the numeric errors tend to be small in the averaged losses.
It would thus make the training insensitive to the subtle differences between the network outputs and the ground truth maps.
We use the Focal Loss (FL), derived from the Focal Loss for object detection \cite{Lin2017FocalDetection}, to perform the hard-voxel-mining with the AH-Net.
We introduce FL regarding the L2 loss that we use in our first DBT image experiment. 
The cross-entropy form of FL that we use in the second CT image experiment can be found in \cite{Lin2017FocalDetection}.
Assuming the L2 loss $D(y,\hat{y})$ is used for supervisely learning a regression map,
\begin{equation}
F_{FL}(y,\hat{y})=(\frac{ln D(y,\hat{y})}{ln D_{max}(y,\hat{y})})^\gamma D(y,\hat{y})
\end{equation}
where $D_{max}(y,\hat{y})$ is the maximum numeric value expected for the L2 loss. The focusing parameter $\gamma$ down-weights the easy voxels. A large $\gamma$ value would make the training focus more on the large numeric errors generated on the hard-voxels. We replace the original L2 loss with FL after a few epochs when the L2 loss barely decreases. 
The training loss could keep descending for more epochs under FL with the output details progressively enhanced.


\section{Experimental Results}

To demonstrate the efficacy and efficiency of the proposed 3D AH-net, we conduct two experiments, namely lesion detection from a Digital Breast Tomosynthesis (DBT) volume and liver tumor segmentation from a Computed Tomography (CT) volume. 
We use ADAM \cite{Kingma2014Adam:Optimization} to optimise all the compared networks with $\beta_1=0.9$, $\beta_2=0.999$ and $\epsilon=10^{-8}$. We use the initial learning-rate $0.0005$ to fine-tune the 2D Multi-Channel GCN.
Then, the learning rate is increased to $0.001$ to fine-tune the AH-Net after the 2D network is transferred. 
We find that 3D networks need a larger learning-rate to converge within a reasonable amount of time. 
All the networks are implemented in Pytorch (\url{http://pytorch.org}).

\subsection{Breast lesion detection from DBT}
We use an in-house database containing 2809 3D DBT volumes acquired from 12 different sites globally. DBT is an advanced form of mammography, which uses low-dose X-Rays to image the breast. Different from 2D mammography that superimposes 3D information into one 2D image, DBT creates 3D pictures of the breast tissue and hence allows radiologists to read these pictures and detect breast cancer more easily, especially in dense breast tissues. The $xy$ plane of DBT images has a high spatial resolution of $0.085mm \times 0.085 mm$ which is much larger than the $z$-dimension of $1mm$. The structures in the $z$-dimension are not only is compressed during the imaging process, but the 3D volumetric information also has large variations due to imaging artefacts.

We have experienced radiologists annotate and validate the lesions in DBT volumes, which might contain zero to several lesions. Each lesion is approximately annotated with a 3D bounding box.
To train the proposed networks as lesion detection networks, we generate 3D multi-variant Gaussian heatmaps that have the same sizes as the original images as
\begin{equation}
f(\bf{p}) = \sum_{\bf{\mu_i} , \bf{\Sigma_i}} \frac{exp(-\frac{1}{2} (\bf{p}-\bf{\mu_i})^T \bf{\Sigma_i} (\bf{p}-\bf{\mu_i}))}{\sqrt{ det(2\pi \bf{\Sigma_i}) }}
\end{equation}
where $\bf{p}$ is a 3D coordinate $x,y,z$; $\bf{\mu_i}$ is the center coordinate of each lesion 3D bounding box; $\bf{\Sigma_i}$ is the covariant matrix of the $i$-th Gaussian determined by the height, width and depth of the 3D bounding box.
Please note that we do not directly predict the bounding box coordinates as the general object detection methods such as Faster RCNN \cite{Ren2015FasterNetworks} because it is sometimes challenging to define the exact boundary of a breast lesion. Also, the voxel-wise confidence maps of lesion presence could be more helpful for clinical decision support than bounding boxes.

\begin{table}[]
\centering
\begin{tabular}{l|lll}
      & \#Volumes & \#Positives & \#Lesions \\ \hline
Train & 2678     & 1111        & 1375      \\
Test  & 131      & 58          & 72
\end{tabular}
\caption{The numbers of volumes (\#Volumes), lesion-positive volumes (\#Positive) and lesions (\#Lesions) in the evaluated DBT dataset.}
\label{tab:split}
\end{table}

We randomly split the database into the training and the testing sets as described in Table.~\ref{tab:split}. 
A volume or a 3D patch is considered positive if at least one lesion is annotated by the radiologist.
We ensure the images from the same patient could only be found either in the training or the testing set.
For training, we extract $256\times256\times32$ 3D patches. 
70\% of the training patches are sampled as positives with at least one lesion included, considering the balance between the voxels within and without a breast lesion. 
The patches are sampled online asynchronously with the network training to form the mini-batches.

Along with the proposed networks, we also train 2D and 3D U-Nets with the identical architecture and parameters \cite{Ronneberger2015U-Net:Segmentation,Cicek20163DAnnotation} as two base-line comparisons.
The 2D U-Net is also trained with input having three input channels.
The 3D U-Net is trained with the same patch sampling strategies as the AH-Net.
All the networks are trained till convergence then the L2 loss function is replaced with the Focal Loss described in Section~\ref{sec:train} for hard-voxel mining. 
The number of convolutional layers and parameters is shown in Table.~\ref{tab:net-stats}.
Using 2D networks, such as the MC-GCN and the 2D U-Net, to process 3D volumes involves repeatedly feeding duplicated images slices.
Thus, they could be slower than the 3D networks when they are used for processing 3D volumes.
We measure the GPU inference time of four networks by forwarding a 3D DBT volume of size $384\times256\times64$ 1000 times on an NVIDIA GTX 1080Ti GPU.
The time spent on operations such as volume slicing is not included in the timing.
The mean GPU time ($ms$) is shown in Table.~\ref{tab:inftime}. 
The GPU inference of AH-Net is 43 times faster than MC-GCN though AH-Net has more parameters.
The speed gain could be brought mostly by avoiding repetitive convolutions on the same slices required by multi-channel 2D networks.

\begin{table}[]
\centering
\begin{tabular}{@{}llr@{}}
\toprule
Network & \#Conv Layers & \#Parameters \\ \midrule
2D-UNet & 15                     & 28,254,528   \\
3D-UNet & 15                     & 5,298,768    \\
*ResNet50 & 53                   & 23,507,904   \\
GCN  & 94                        & 23,576,758   \\
AH-Net  & 123                    & 27,085,500   \\ \bottomrule
\end{tabular}
\caption{The number of convolutional layers (\#Conv Layers) and model float parameters (\#Parameters) respectively in 2D-UNet, 3D-UNet, ResNet50, GCN and AH-Net. ResNet50 is shown here as a reference to be compared with GCN with a simple decoder added.}
\label{tab:net-stats}
\end{table}

\begin{table}[]
\centering
\begin{tabular}{@{}lllll@{}}
\toprule
  & 2D U-Net & 3D U-Net & MC-GCN  & 3D AH-Net \\ \midrule
$ms$ & 699.3     & 2.3   & 775.2 & 17.7     \\ \bottomrule
\end{tabular}
\caption{The GPU inference time ($ms$) of different networks on a $384\times256\times64$ volume computed by averaging 1000 inferences with a NVIDIA GTX 1080Ti.}
\label{tab:inftime}
\end{table}

\begin{figure}[!htb]
  \centering
  \includegraphics[width=1\linewidth]{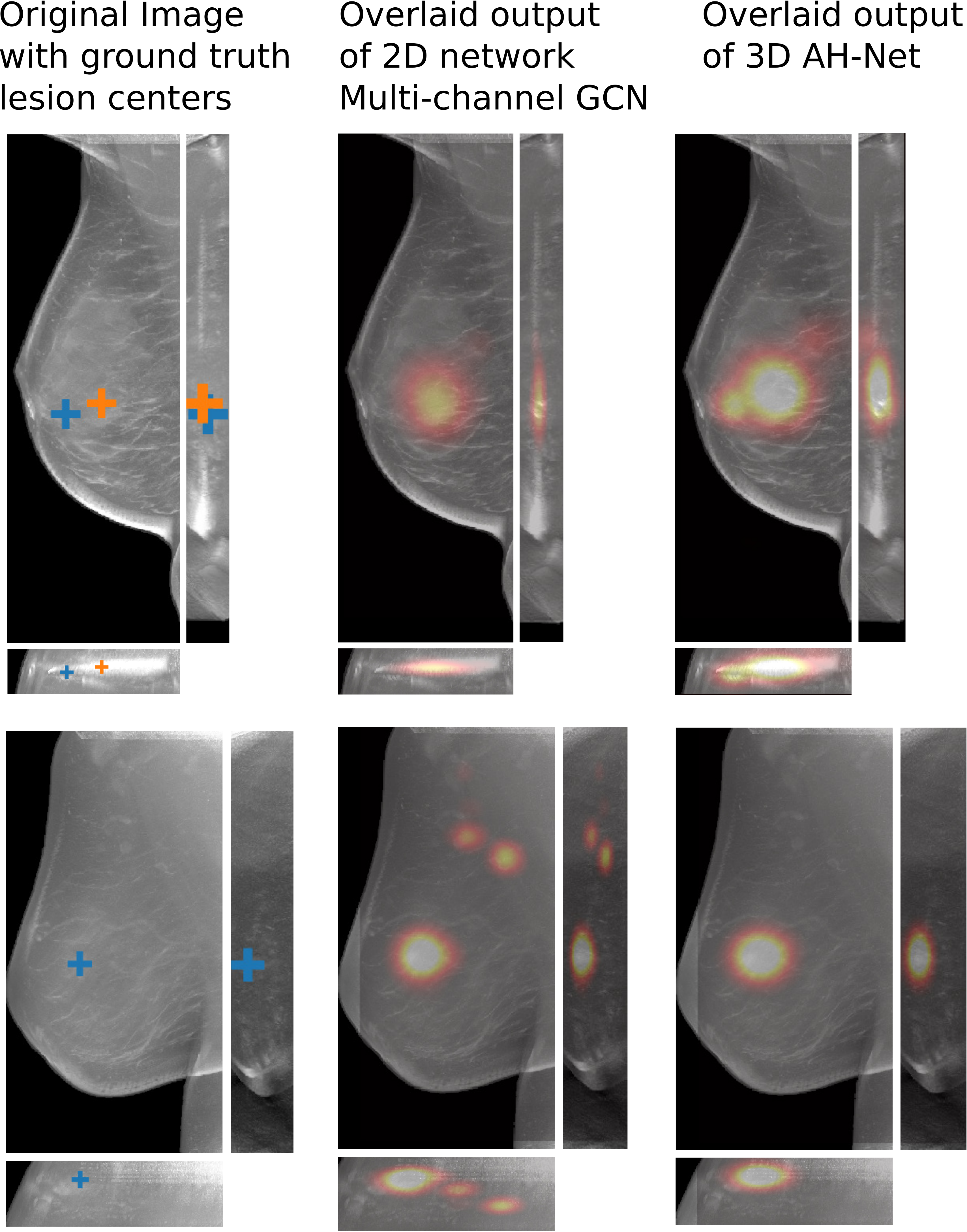}
  \caption{The visual comparisons of the network responses on 2 different DBT volumes from 2D GCN and the 3D AH-Net with the encoder weights transferred from it. 
  Each volume is visualized with the maximum intensity projection of the $xy$ plane (top-left), the $xz$ plane (bottom) and the $yz$ plane (right).
  The ground truth lesion centres are shown on the left. With the additional AH-Net Decoders, 3D AH-Net could effectively detect the missing lesion in the first volume (upper row) and remove the false positives in the second volume (lower row).}
  \label{fig:visual}
\end{figure}

\begin{figure}
\centering
\includegraphics[width=1\linewidth]{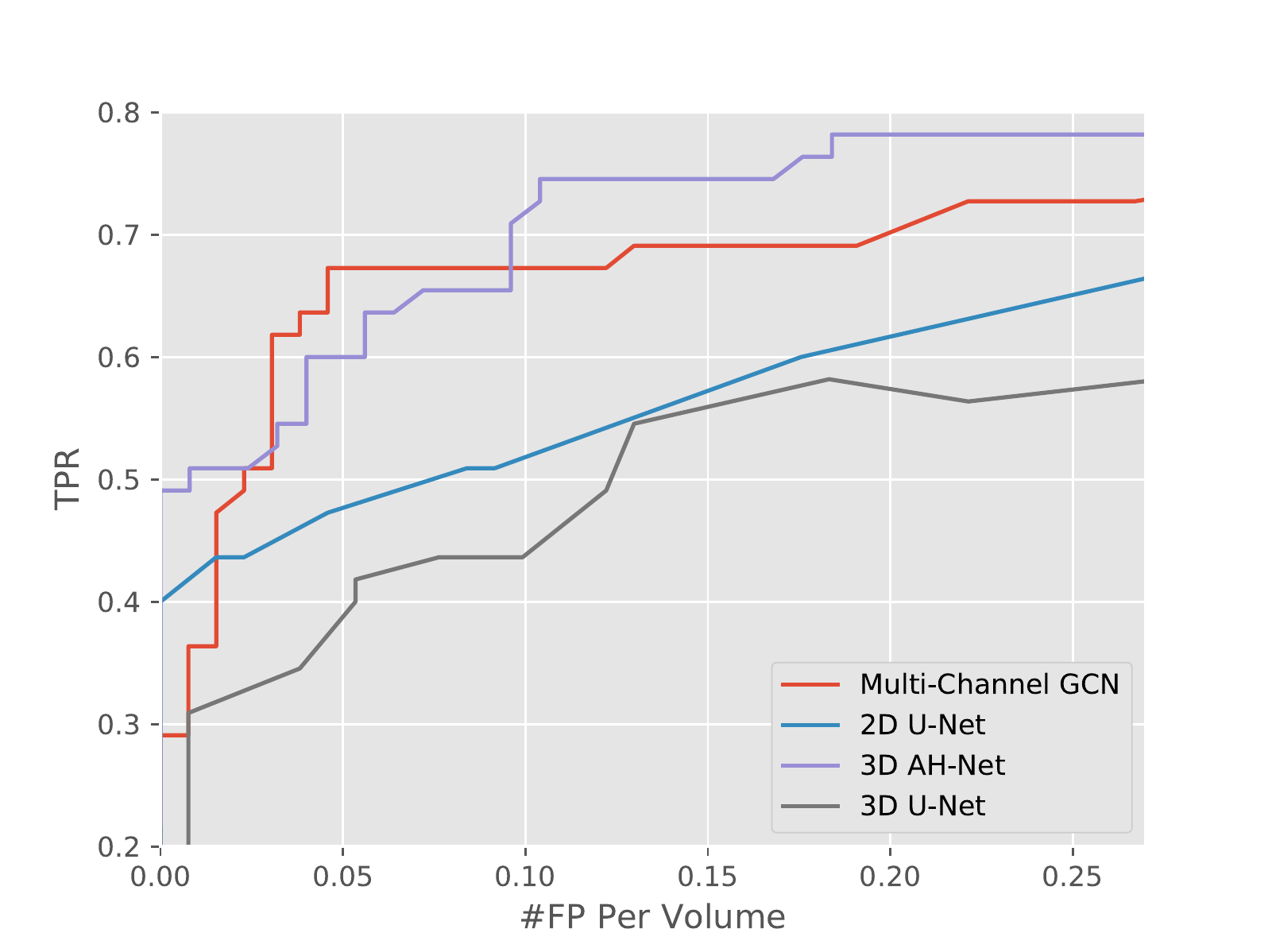}
\caption{The Free Response Operating Characteristic (FROC) curves regarding the lesion detection performance.}
\label{fig:box}
\end{figure}

\begin{table*}[]
\centering
\begin{tabular}{l|llllll}
\hline
         & FP=0.01           & FP=0.05           & FP=0.10           & FP=0.15           & FP=0.20           & FP=0.25 \\ \hline
2D U-Net & 0.4238            & 0.4767            & 0.5181            & 0.5723            & 0.6166            & 0.6506           \\
3D U-Net & 0.2448            & 0.3877            & 0.4381            & 0.5592            & 0.5738            & 0.5733            \\
GCN      & 0.3385            & \textbf{0.6727} & 0.6727            & 0.6909            & 0.7018            & 0.7272\\
AH-Net   & \textbf{0.4931} & 0.6000               & \textbf{0.7272} & \textbf{0.7454} & \textbf{0.7818} & \textbf{0.7818} \\ \hline
\end{tabular}
\caption{The quantitative metrics of the compared networks on the DBT dataset. True positive rate (TPR) sampled at five different numbers of false positive (FP) findings allowed are shown in the first five columns. 
}
\label{tab:roc-numbers}
\end{table*}

Non-maximal suppression is performed on the network output map to obtain the lesion locations.
The network responses at the local maximal voxels are considered as the  confidence scores of the cancerous findings. Fig. \ref{fig:visual} shows some visual comparison of the networks output.

By altering a threshold to filter the response values, we can control the balance between the False Positive Rate (FPR) and True Positive Rate (TPR).
The lesion detected by the network is considered a true positive finding if the maximal point resides in a 3D bounding box annotated by the radiologist. Similarly, if a bounding box contains a maximal point, we consider it is detected by the network. The maximal points are otherwise considered as false positive findings.
We evaluate the lesion detection performance by plotting the Free Response Operating Characteristic (FROC) curves, which measures the True Positive Rate (TPR) against the number of false positive (\#FP) allowed per volume.
TPR represents the percentage of lesions that have been successfully detected by the network.
FPR represents the percentage of lesions that the network predicted that are false positives.
As shown in Fig.\ref{fig:box}, the proposed AH-Net out-performs both the 2D and 3D U-Net with large margins.
Compared to the performance of the 2D network (Multi-Channel GCN), the 3D AH-Net generates higher TPR for a majority of thresholds, except the region around 0.05 per volume false positives. It is noticeable that AH-Net also obtains nearly 50\% TPR even when only 0.01 false positive findings are allowed per volume. 
Interestingly, the performance of 3D-UNet is slightly worse than that of 2D-UNet, though the DBT volumes have three dimensions. 
This might be caused by the anisotropic resolution of DBT images and the limited number of parameters constrained by the GPU memory.
The FROC numbers are summarised in Table.~\ref{tab:roc-numbers}.

\begin{figure*}
\centering
\includegraphics[width=1\linewidth]{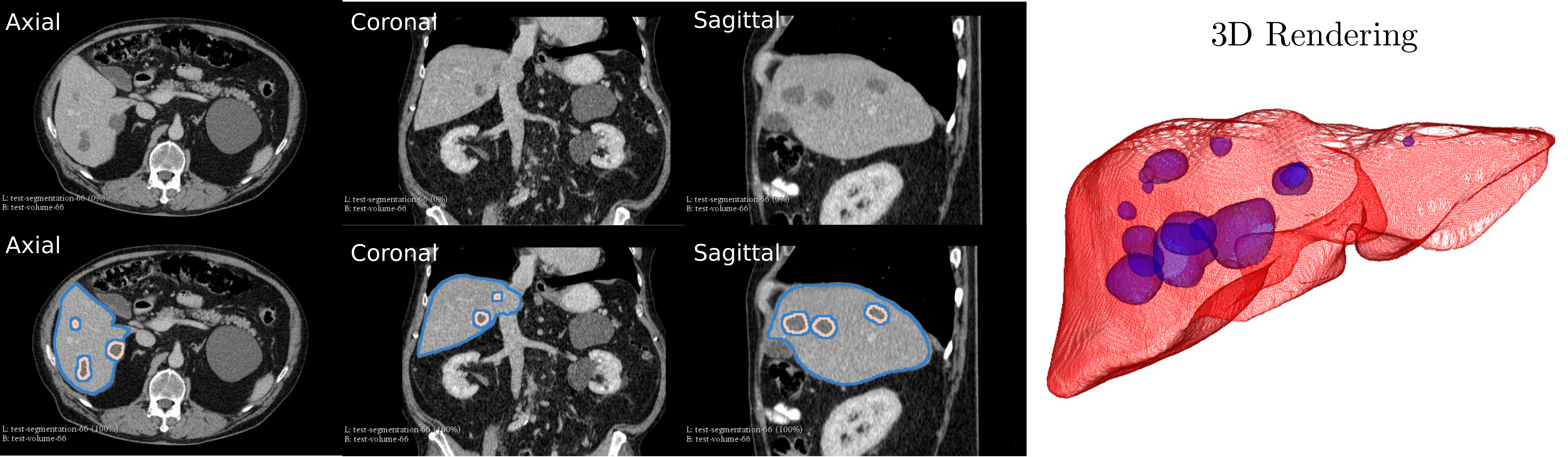}
\caption{The example liver lesion segmentation results from 3D AH-Net. The segmented contours of liver (blue) and liver lesion (pink) are overlaid on 3 slices viewed from different orientations (Axial, Coronal and Sagittal). The segmentations are rendered in 3D on the right.}
\label{fig:liver-slice}
\vspace{-5mm}
\end{figure*}

\subsection{Liver and liver tumor segmentation from CT}
The second evaluation dataset was obtained from the liver lesion segmentation challenge in MICCAI 2017 (\url{lits-challenge.com}), which contains 131 training and 70 testing 3D contrast-enhanced abdominal CT scans. Liver lesion is one of the most commonest cancer worldwide. It is estimated that 28920 people will die of liver lesion and 40710 new cases will be diagnosed in 2017 \cite{AmericanCancerSociety2017Cancer2017}.
Automatic segmentation of liver and lesion is challenging due to the heterogeneous and diffusive appearance of both liver and lesions. Also, the number, shape, location of the lesions varies a lot among different volumes. The data and ground-truth masks were provided by various clinical sites around the world. The ground truth masks contain both liver and lesion labels. Most CT scans consist of anisotropic resolution: the between-slice resolution ranges from 0.45mm to 6.0mm while the within-slice resolution varies from 0.55mm to 1.0mm. All scans cover the abdominal regions but may extend to head and feet. Other than the liver lesion, other diseases may also exist in these data, which further increases the task difficulty.

In preprocessing, the abdominal regions are truncated from the CT scans using the liver center biomarker detected by a reinforcement learning based algorithm \cite{Ghesu2017RobustData}. While this step makes the network concentrate on the targeting region, its accuracy is not critical as we choose a relatively large crop region which usually ranges from the middle of the lung to the top of the pelvis. The image intensity is truncated to the range of [-125,225] HU based on the intensity distribution of liver and lesion in the training data.
Due to the limited number of training data, we applied random rotation (within ${\pm}20$ degree in the $xy$ plane), random scaling (within ${\pm}0.2$ in all directions), and random mirror (within $xy$ plane) to reduce overfitting.

We first train the MC-GCN with pre-trained ResNet50 as the back-bone network. The input size of stacked 2D slices is $512\times512$ with three channels. After convergence, the weights of the encoder part of MC-GCN are transformed to the corresponding layers of a 3D AH-Net, which is then finetuned using 3D patches with size $192\times192\times64$. The weights of other layers are randomly initialized. In the training of both networks, the cross-entropy loss is used at the beginning until convergence, which is then replaced by the Focal Loss for hard voxel mining \cite{Lin2017FocalDetection}.

The performance of AH-Net is listed in Table \ref{tab:liver}, together with other six top-ranked submissions retrieved from the LITS challenge leaderboard.
These submissions employ various types of neural network architectures: 2D, 3D, 2D-3D hybrid, and model fusion.
Two evaluation metrics are adapted: (1) Dice Global (DG) which is the dice score combining all the volumes into one; (2) Dice per Case (DPC) which averages of the dice scores of every single case. The Dice score between two masks is defined as $DICE(A,B) = 2|A\cap B|/(|A|+|B|)$. Our results achieve the state-of-the-art performance in three of the four metrics, including the dice global score of the lesions, dice global and dice per case score of the livers, which proves the effectiveness of AH-Net for segmenting 3D images with diverse anisotropic resolution. Some example results are shown in Fig.\ref{fig:liver-slice}.


\begin{table}[]
\centering
\begin{tabular}{@{\extracolsep{4pt}}lllll@{}}
\hline
           & \multicolumn{2}{l}{\bfseries Lesion} & \multicolumn{2}{l}{\bfseries Liver} \\ \cline{2-3}\cline{4-5}
Method     & DG               & DPC              & DG              & DPC              \\ \hline
leHealth   &  0.794            &  \bf{0.702}           & 0.964            & 0.961        \\
H-DenseNet \cite{Li2017H-DenseUNet:Volumes} & 0.829          &    0.686         & 0.965            &   0.961          \\
hans.meine &   0.796           & 0.676            & 0.963            &  0.960           \\
medical    & 0.783             & 0.661            & 0.951            & 0.951            \\
deepX      & 0.820             & 0.657            & 0.967            & \bf{0.963}            \\
superAI    & 0.814             & 0.674            & -                & -                \\ \hline
GCN        & 0.788             & 0.593                 & 0.963                 & 0.951 \\
3D AH-Net  & \bf{0.834}             & 0.634            & \bf{0.970}            & \bf{0.963}            \\ \hline
\end{tabular}
\caption{The liver lesion segmentation (LITS) challenge results with the dice global (DG) and dice per case (DPC). The compared results were obtained from the LITS challenge leaderboard (\url{lits-challenge.com/\#results}).}
\vspace{-5mm}
\label{tab:liver}
\end{table}

\section{Conclusion}
In this paper, we propose the 3D Anisotropic Hybrid Network (3D AH-Net) which is capable of transferring the convolutional features of 2D images to 3D volumes with anisotropic resolution.
By evaluating the proposed methods on both a large-scale in-house DBT dataset and a highly competitive open challenge dataset of CT segmentation, we show our network could obtain the state-of-the-art results.
AH-Net generalizes better than the traditional 3D networks, such as 3D U-Net \cite{Cicek20163DAnnotation} due to the features transferred from a 2D network and the anisotropic convolution blocks.
The GPU inference of AH-Net is also much faster than piling the results from a 2D network.
Though AH-Net is designed for anisotropic volumes, we believe it could also be applied to volumes with resolution closed to being isotropic, such as CT and MRI.

\noindent\textbf{Disclaimer}: This feature is based on research, and is not commercially available. Due to regulatory reasons, its future availability cannot be guaranteed.

{\small
\bibliographystyle{ieee}
\bibliography{mendeley}
}

\appendices

\section{Visual Cases of the DBT dataset}
\begin{figure}[!htb]
\includegraphics[width=1\linewidth]{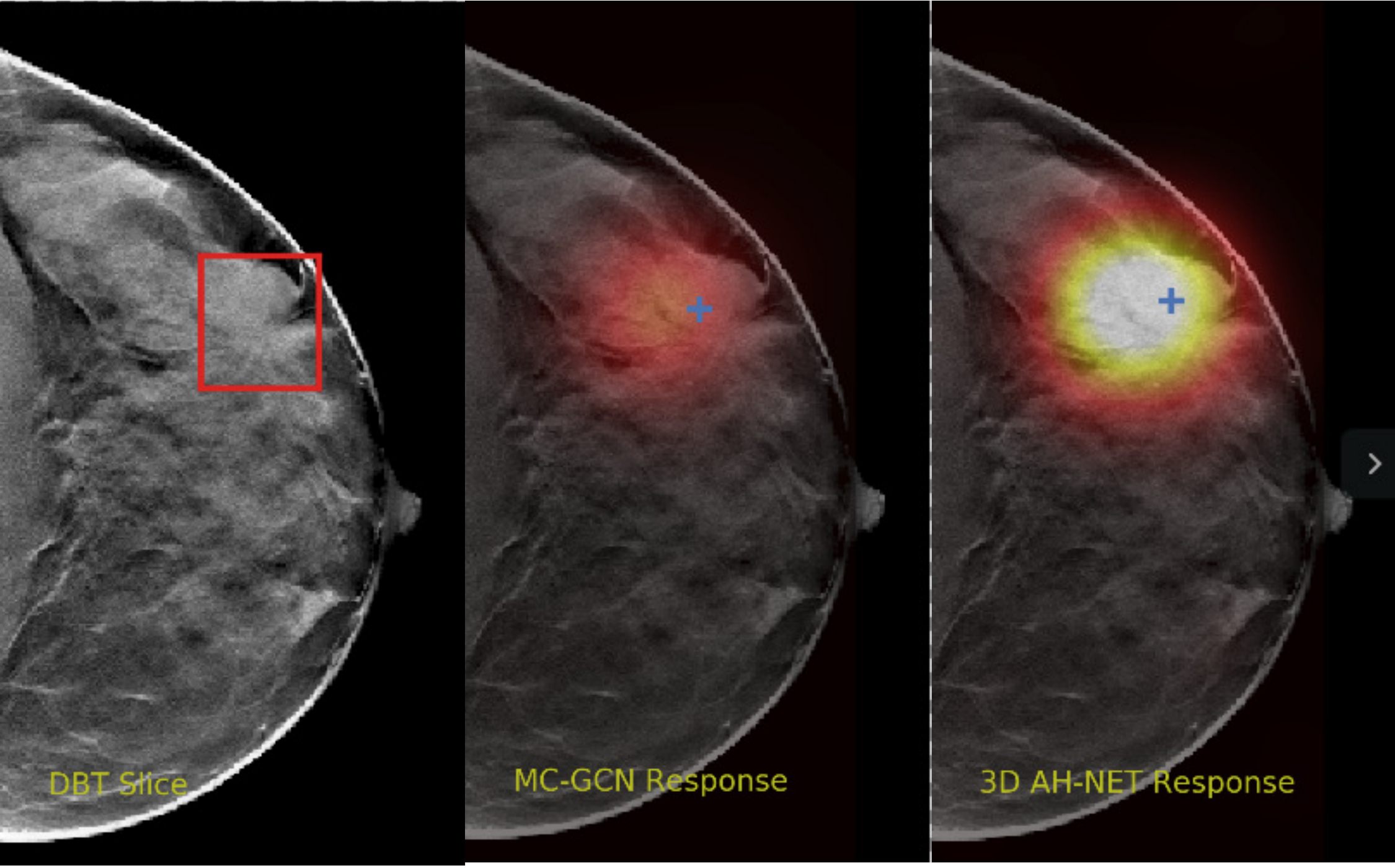}
\caption{Example DBT slice 1 with a lesion that can be detected by both MC-GCN and 3D AH-Net. Though the lesion is blended in the dense breast tissues, our network is able to detect it according to the speculations around the lesion boundary.}
\label{fig:DBT-A1}
\end{figure}

\begin{figure}[!htb]
\includegraphics[width=1\linewidth]{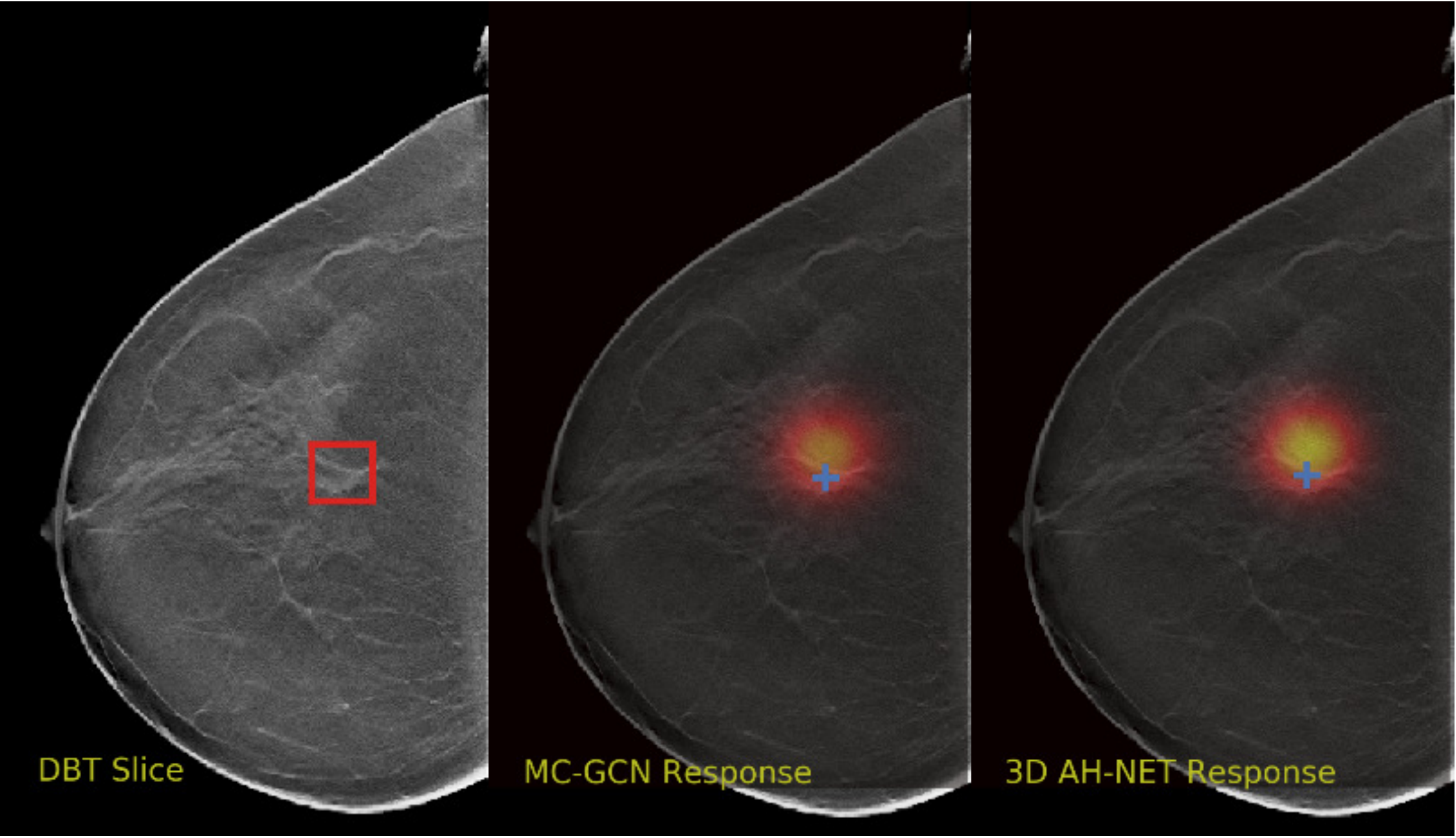}
\caption{Example DBT slice 2 with a lesion that can be detected by both MC-GCN and 3D AH-Net. The lesion is small and can also be identified with the architectural distortion in the surrounding tissues.}
\label{fig:DBT-A2}
\end{figure}

\begin{figure}[!htb]
\includegraphics[width=1\linewidth]{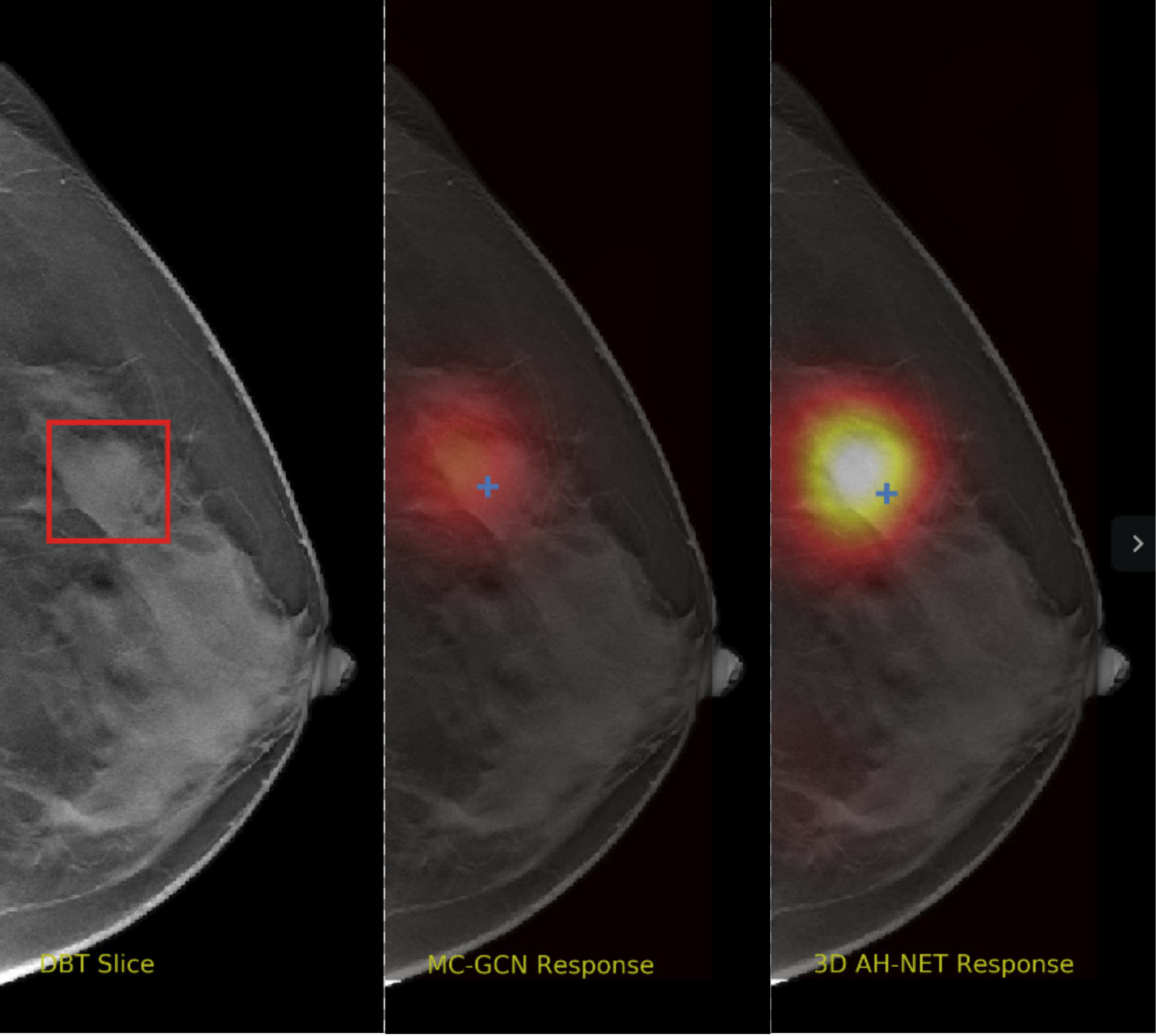}
\caption{Example DBT slice 3 with a lesion that can be detected by both MC-GCN and 3D AH-Net. The lesion is blended in the dense breast tissues.}
\label{fig:DBT-A3}
\end{figure}

\begin{figure}[!htb]
\includegraphics[width=1\linewidth]{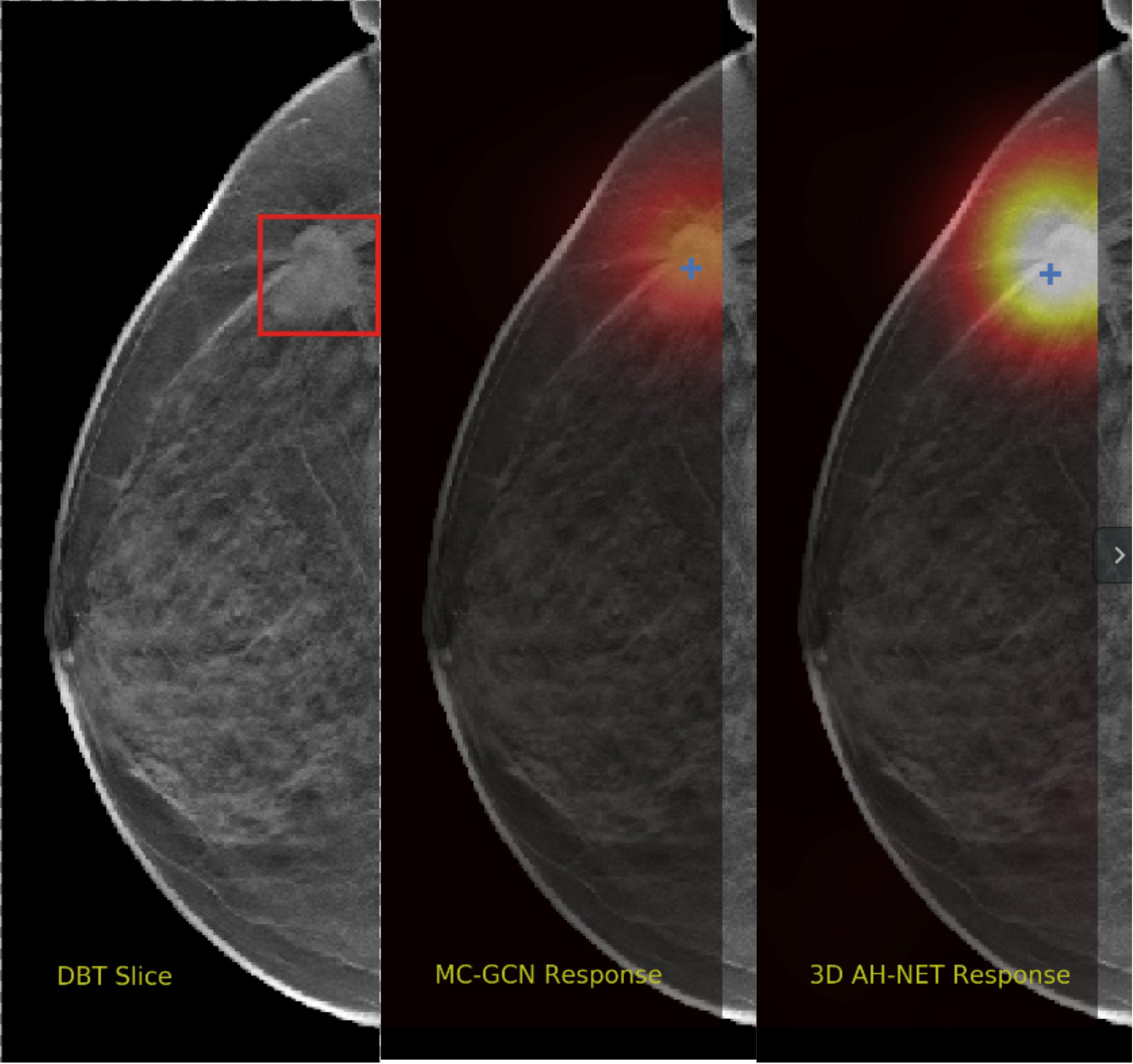}
\caption{Example DBT slice 4 with a lesion that can be detected by both MC-GCN and 3D AH-Net. The lesion has clear boundaries and speculations.}
\label{fig:DBT-A4}
\end{figure}

\begin{figure}[!htb]
\includegraphics[width=1\linewidth]{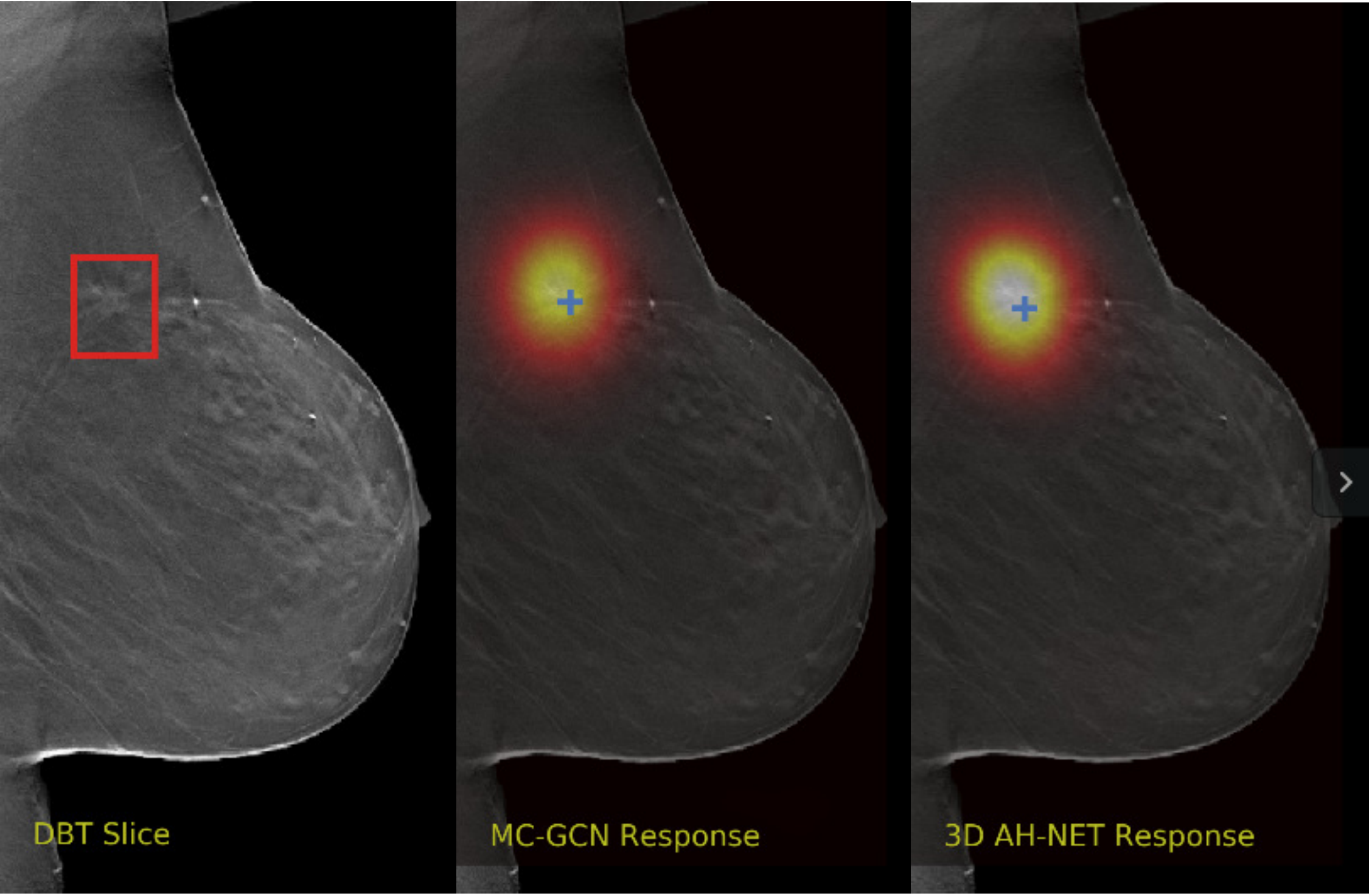}
\caption{Example DBT slice 5 with a lesion that can be detected by both MC-GCN and 3D AH-Net. The small lesion causes architectural distortion in the surrounding tissues.}
\label{fig:DBT-A5}
\end{figure}

\begin{figure}[!htb]
\includegraphics[width=1\linewidth]{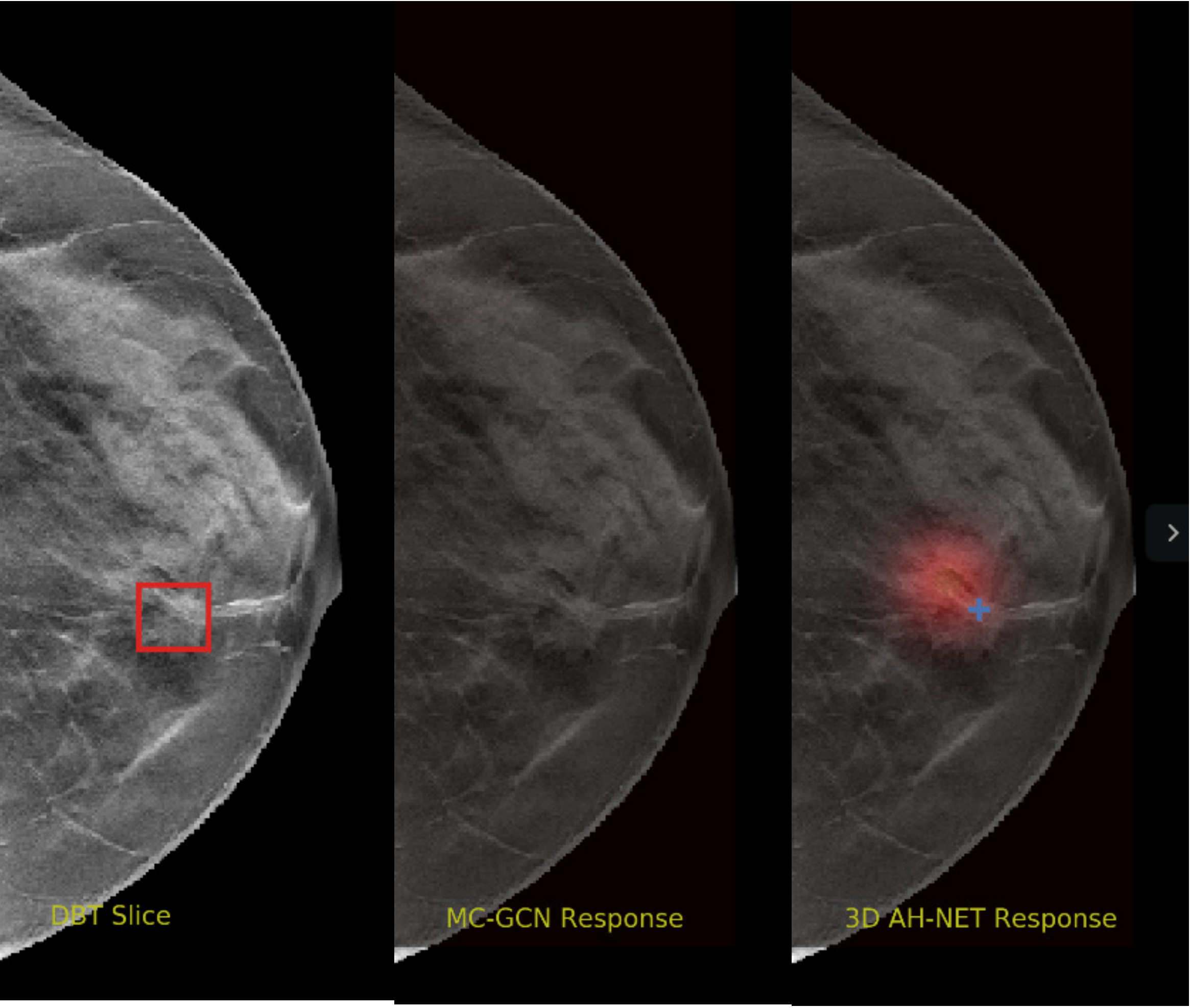}
\caption{Example DBT slice 6 with a lesion that can only be detected by 3D AH-Net. The lesion is highly blended within the dense breast tissues which makes it challenging for both the radiologists and the networks to detect through a single slice. In contrast, the lesion can be detected by considering the consistency of the structure across a few neighbouring slices.}
\label{fig:DBT-B1}
\end{figure}

\begin{figure}[!htb]
\includegraphics[width=1\linewidth]{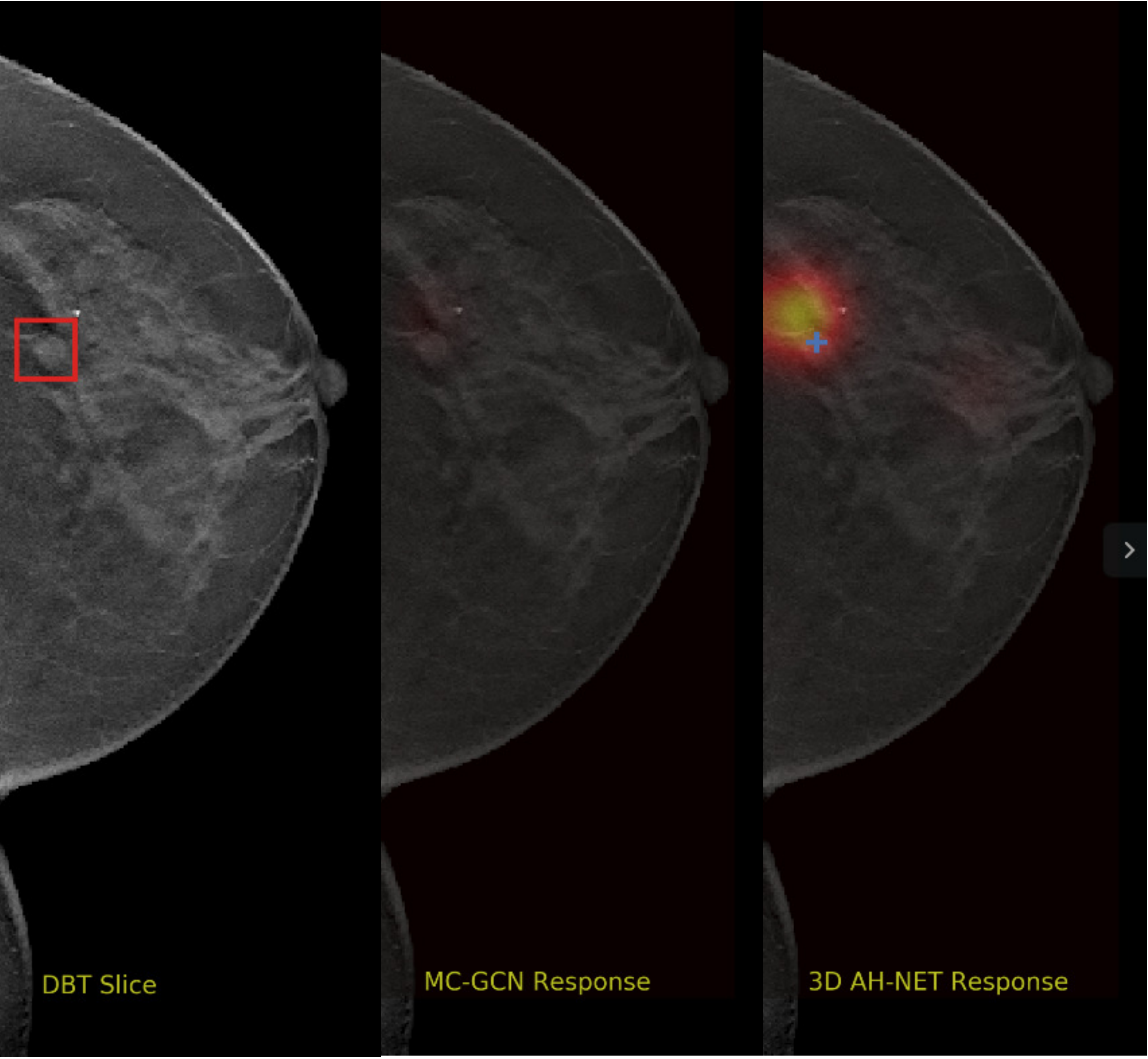}
\caption{Example DBT slice 7 with a lesion that can only be detected by 3D AH-Net. The lesion is highly blended within the dense breast tissues which makes it challenging for both the radiologists and the networks to detect through a single slice. In contrast, the lesion can be detected by considering the consistency of the structure across a few neighbouring slices.}
\label{fig:DBT-B2}
\end{figure}

\begin{figure}[!htb]
\includegraphics[width=1\linewidth]{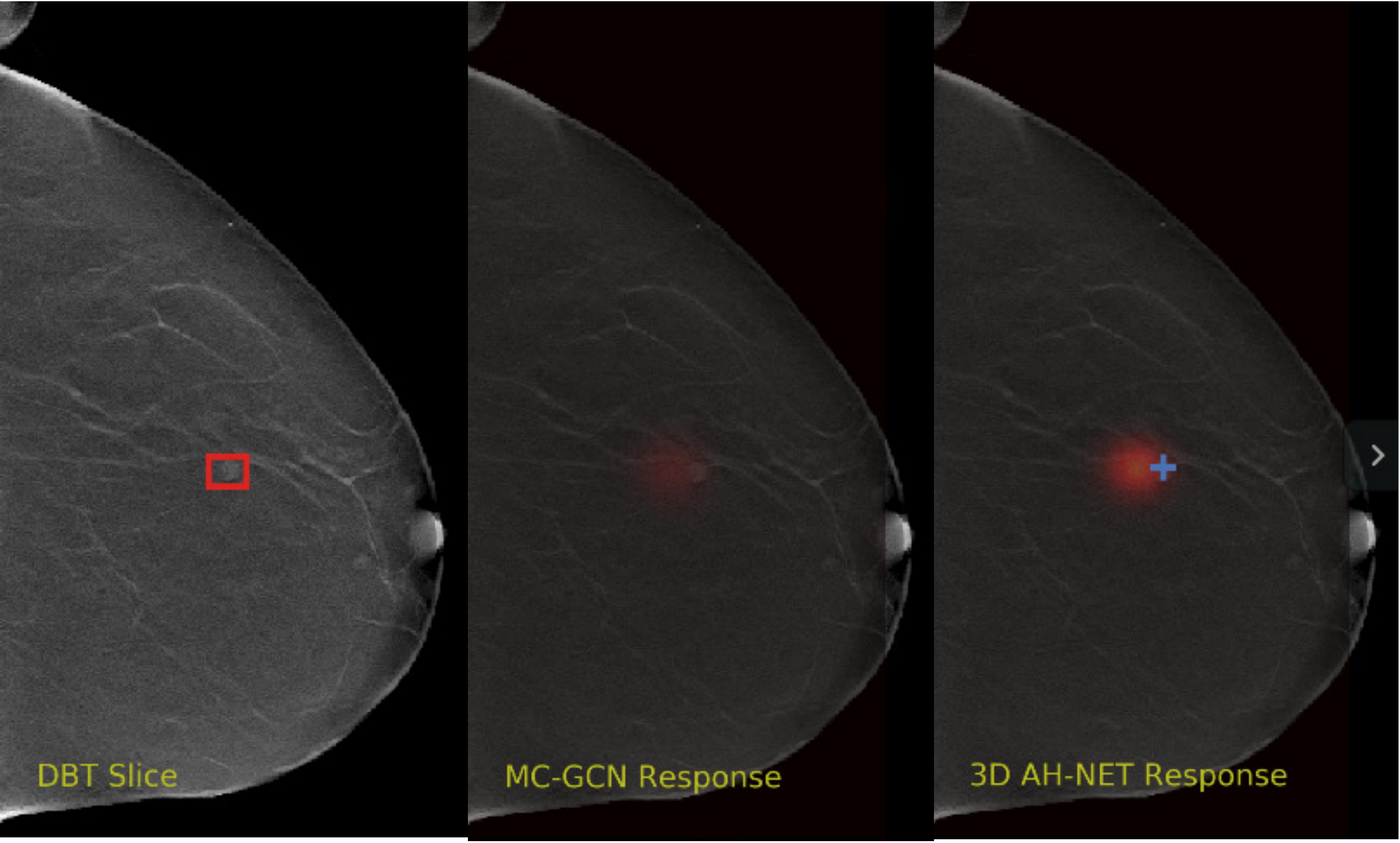}
\caption{Example DBT slice 8 with a lesion that can only be detected by 3D AH-Net. The lesion is small and hard to be distinguished from other breast tissues. The lesion can be detected by considering the consistency of the structure across a few neighbouring slices.}
\label{fig:DBT-B3}
\end{figure}

\begin{figure}[!htb]
\includegraphics[width=1\linewidth]{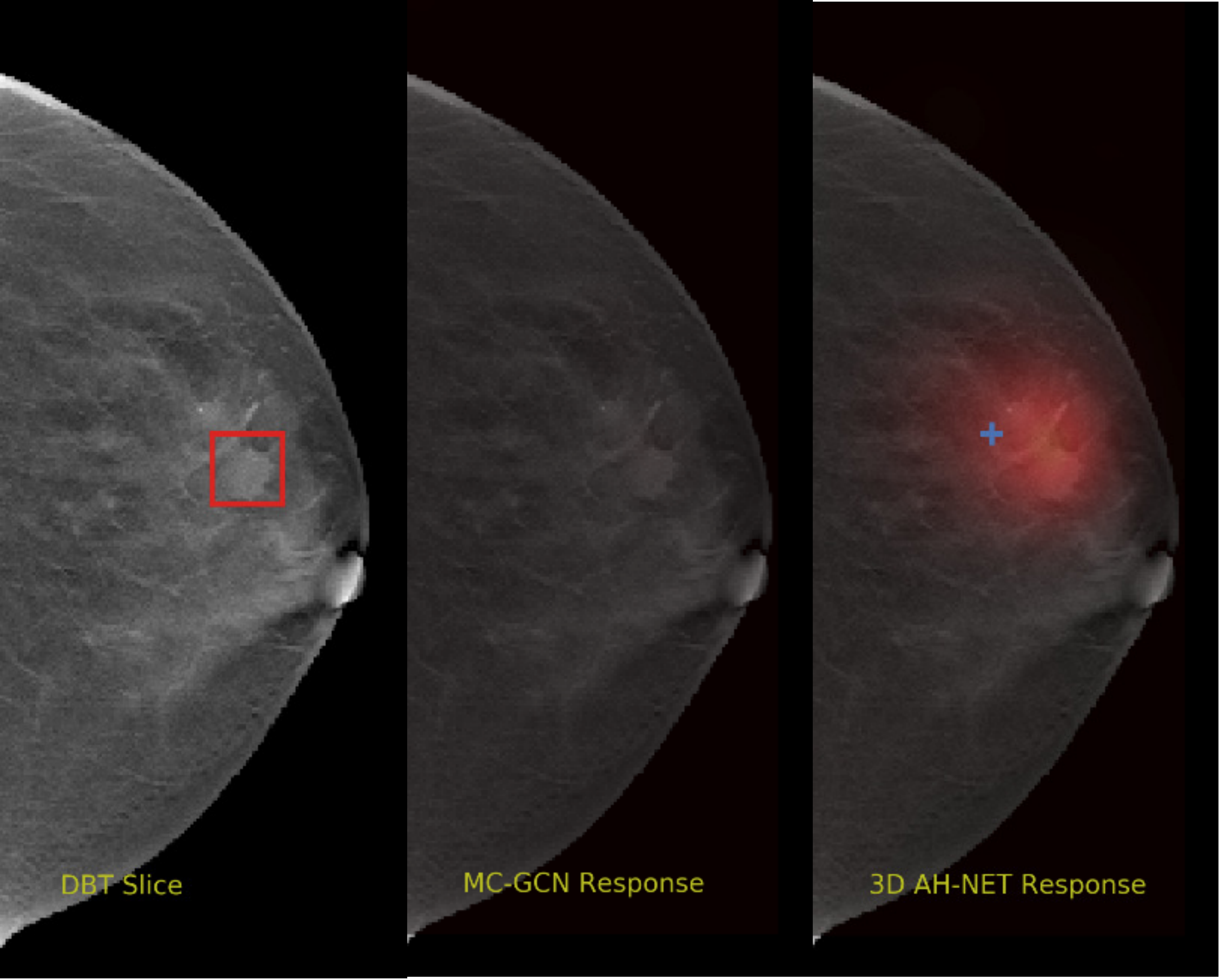}
\caption{Example DBT slice 9 with a lesion that can only be detected by 3D AH-Net. The lesion is highly blended within the dense breast tissues which makes it challenging for both the radiologists and the networks to detect with only a 2D view of the structure. The lesion can be detected by considering the consistency of the structure across a few neighbouring slices.}
\label{fig:DBT-B4}
\end{figure}

\begin{figure}[!htb]
\includegraphics[width=1\linewidth]{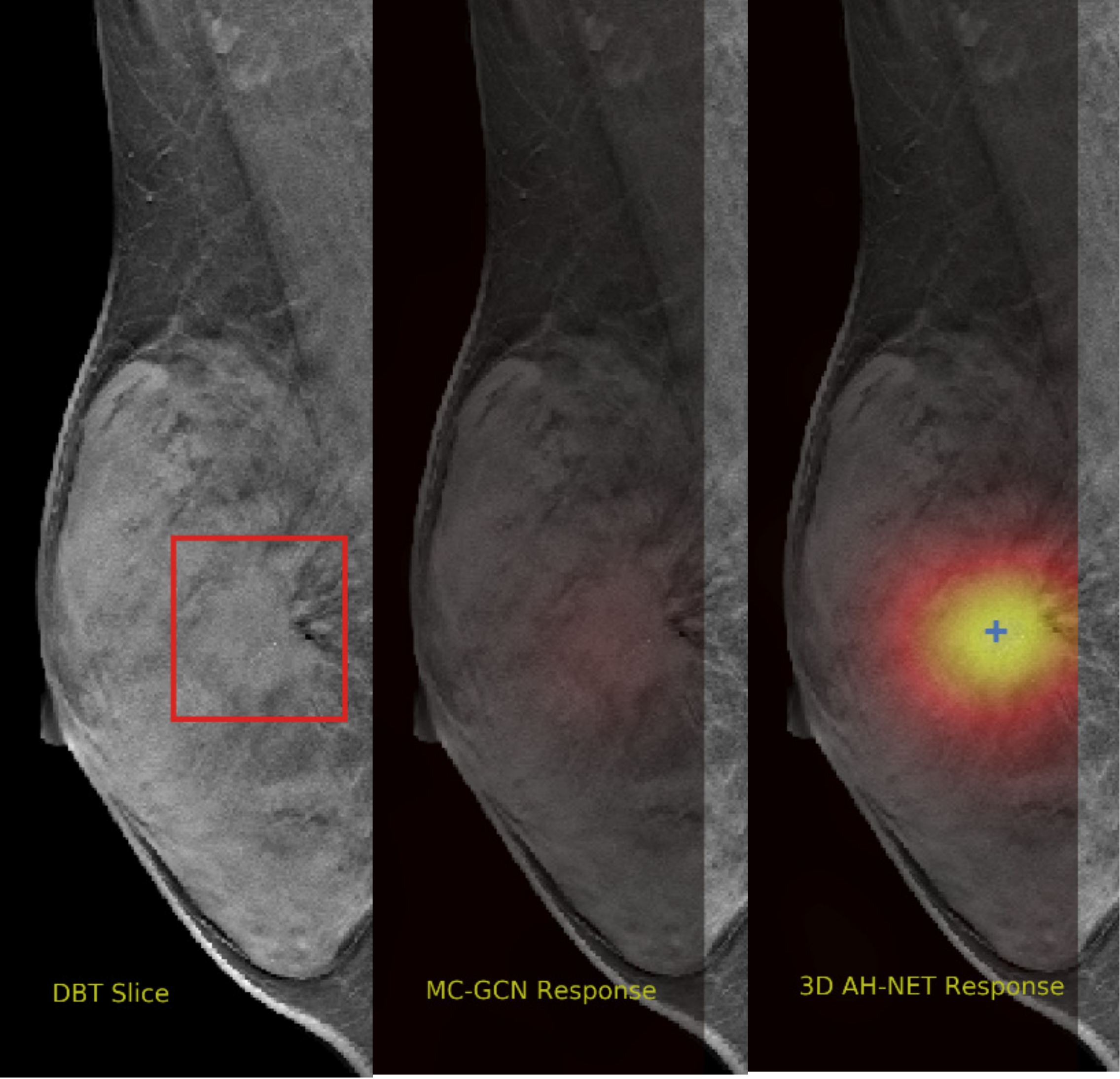}
\caption{Example DBT slice 10 with a lesion that can only be detected by 3D AH-Net. The lesion is highly blended within the dense breast tissues which makes it challenging for both the radiologists and the networks to detect through a slice 2D slice. The lesion can be detected by considering the consistency of the structure across a few neighbouring slices.}
\label{fig:DBT-B5}
\end{figure}

\begin{figure}[!htb]
\includegraphics[width=1\linewidth]{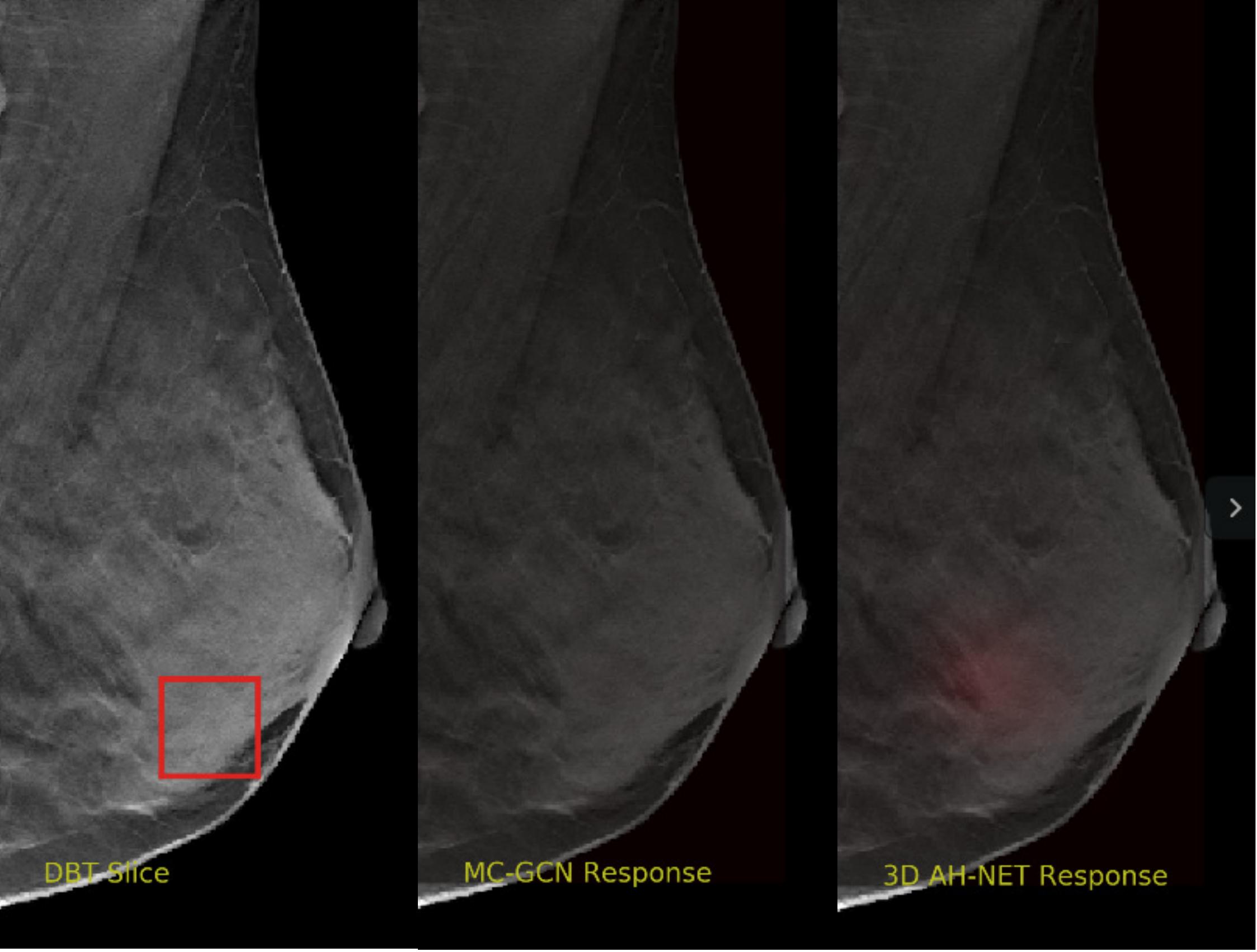}
\caption{Example DBT slice 11 with a lesion that neither network is able to detect. The contrast between lesion and the normal tissue is too low.}
\label{fig:DBT-C1}
\end{figure}

\begin{figure}[!htb]
\includegraphics[width=1\linewidth]{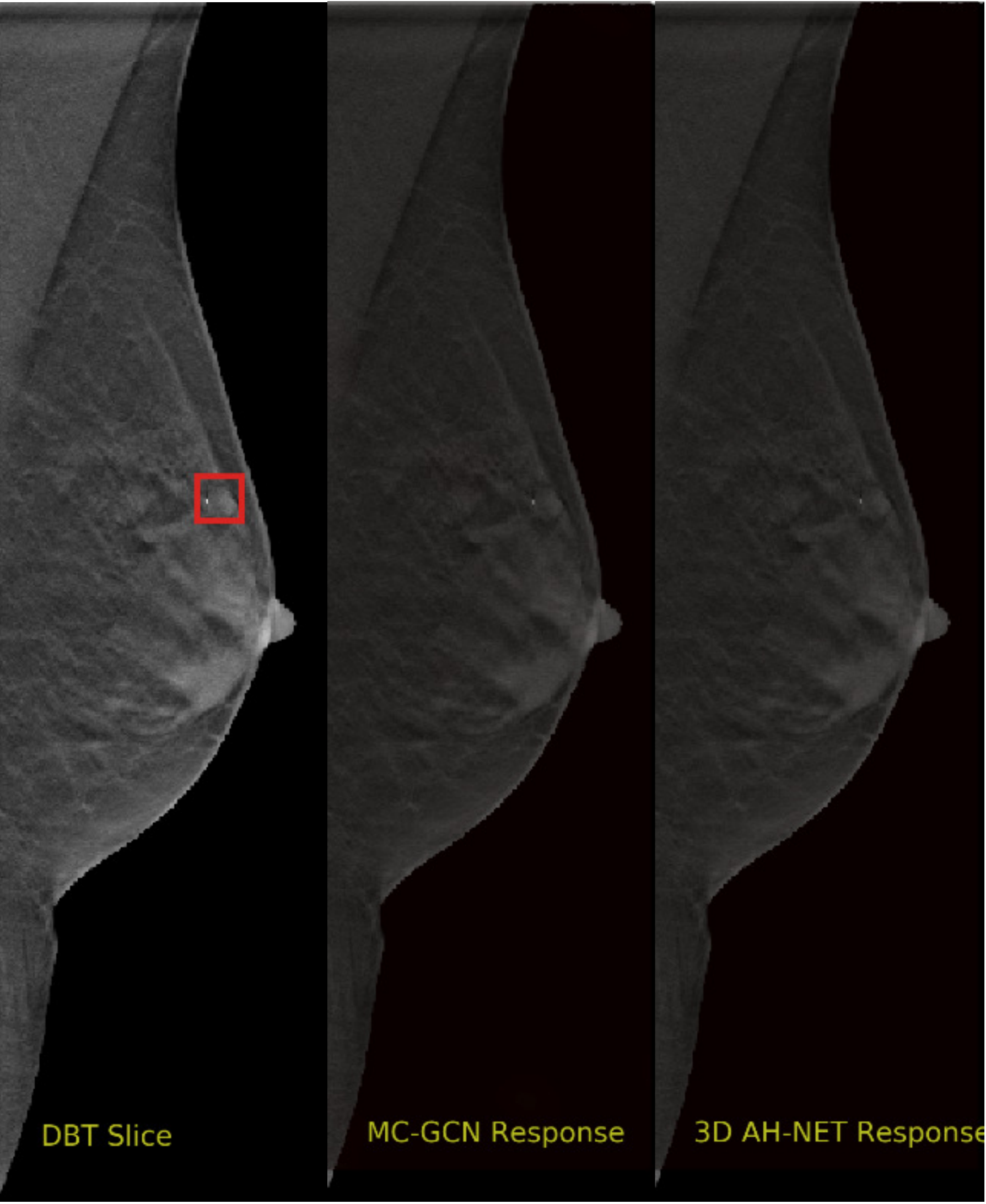}
\caption{Example DBT slice 12 with a lesion that neither network is able to detect. The contrast between lesion and the normal tissue is too low.}
\label{fig:DBT-C2}
\end{figure}

\begin{figure}[!htb]
\includegraphics[width=1\linewidth]{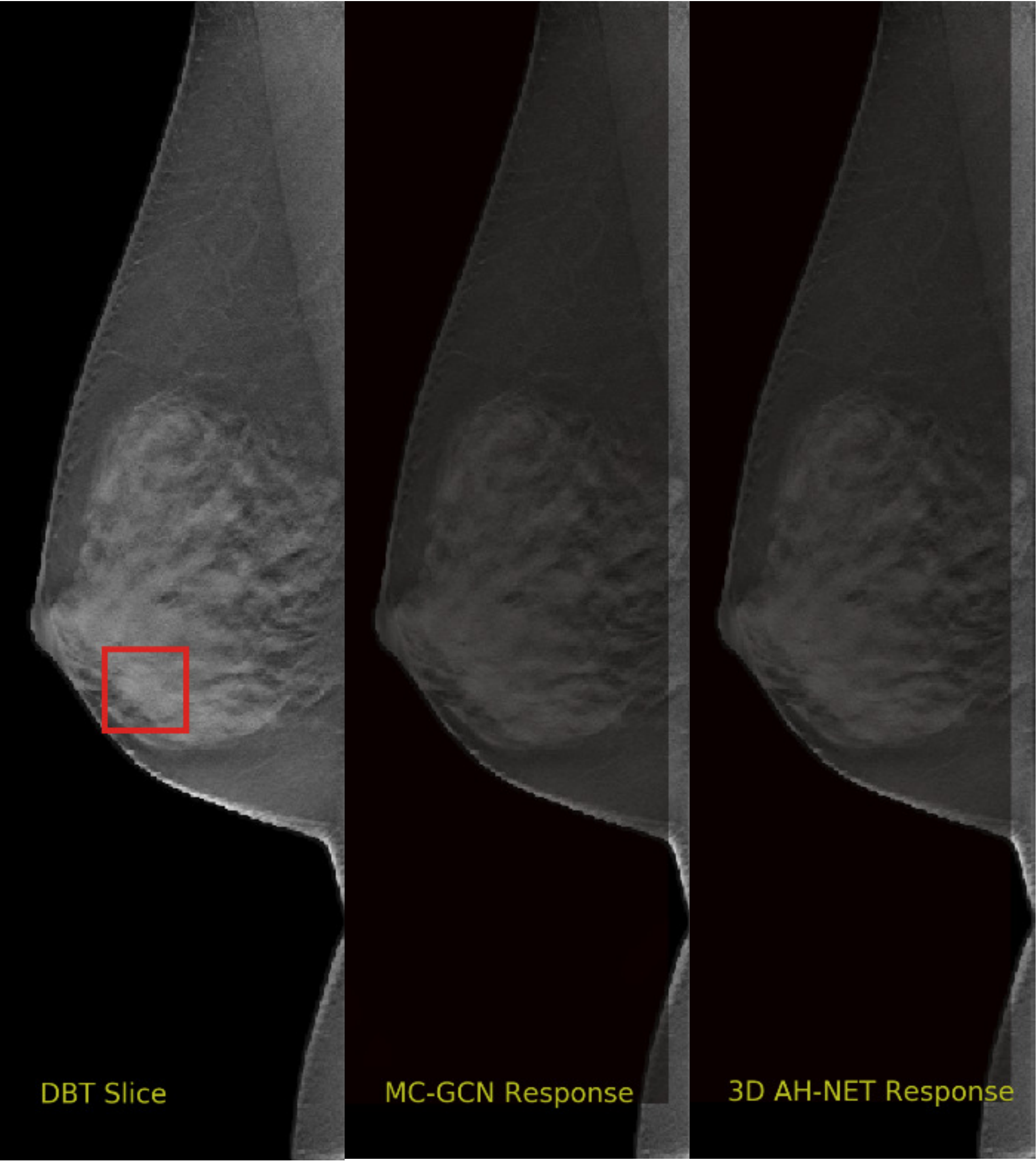}
\caption{Example DBT slice 13 with a lesion that neither network is able to detect. The contrast between lesion and the normal tissue is too low.}
\label{fig:DBT-C3}
\end{figure}

\begin{figure}[!htb]
\includegraphics[width=1\linewidth]{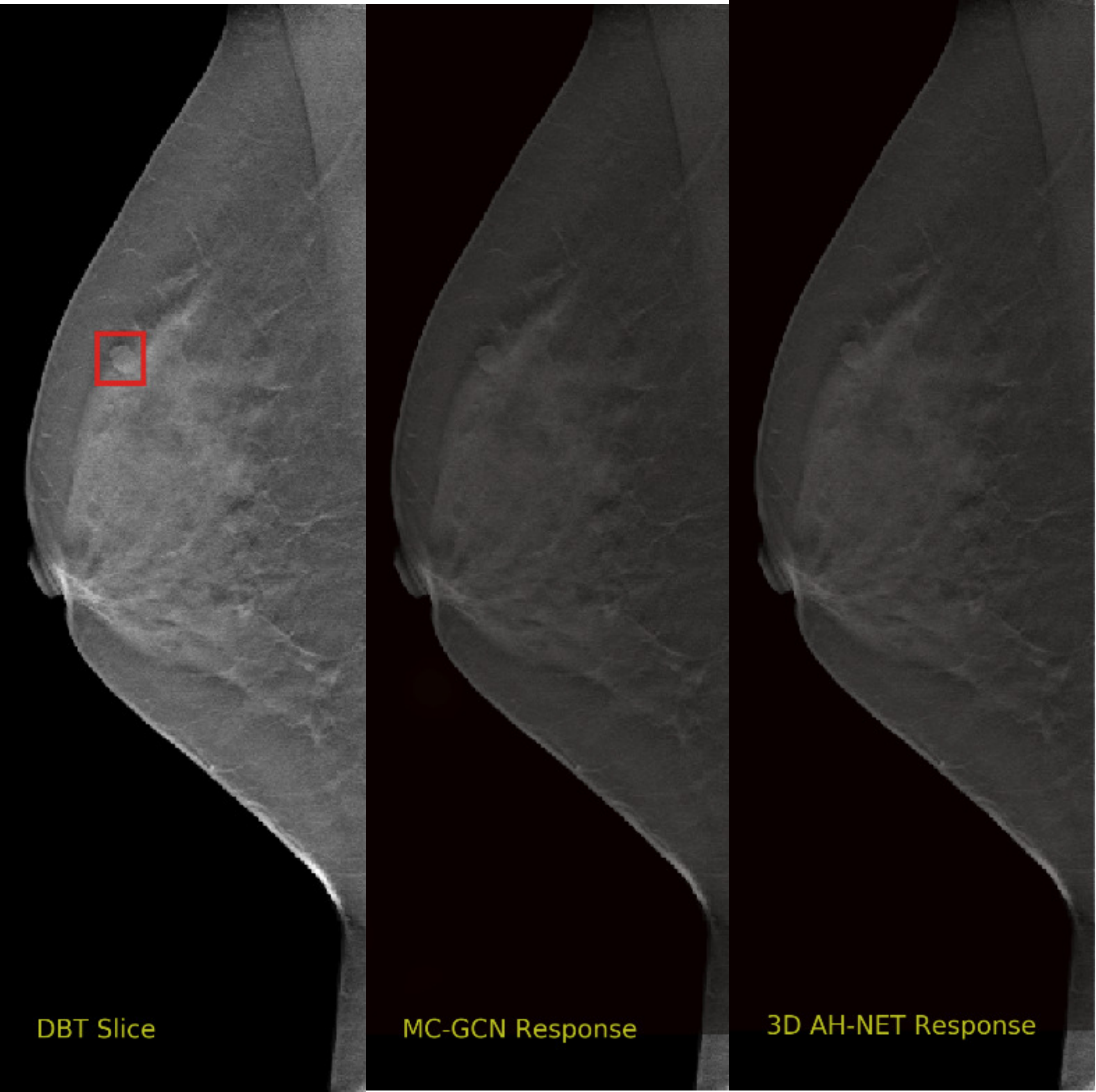}
\caption{Example DBT slice 14 with a lesion that neither network is able to detect. Although the lesion has a roundish shape, it is hard for the network to distinguish them from the lymph nodes or skin moles.}
\label{fig:DBT-C4}
\end{figure}

\begin{figure}[!htb]
\includegraphics[width=1\linewidth]{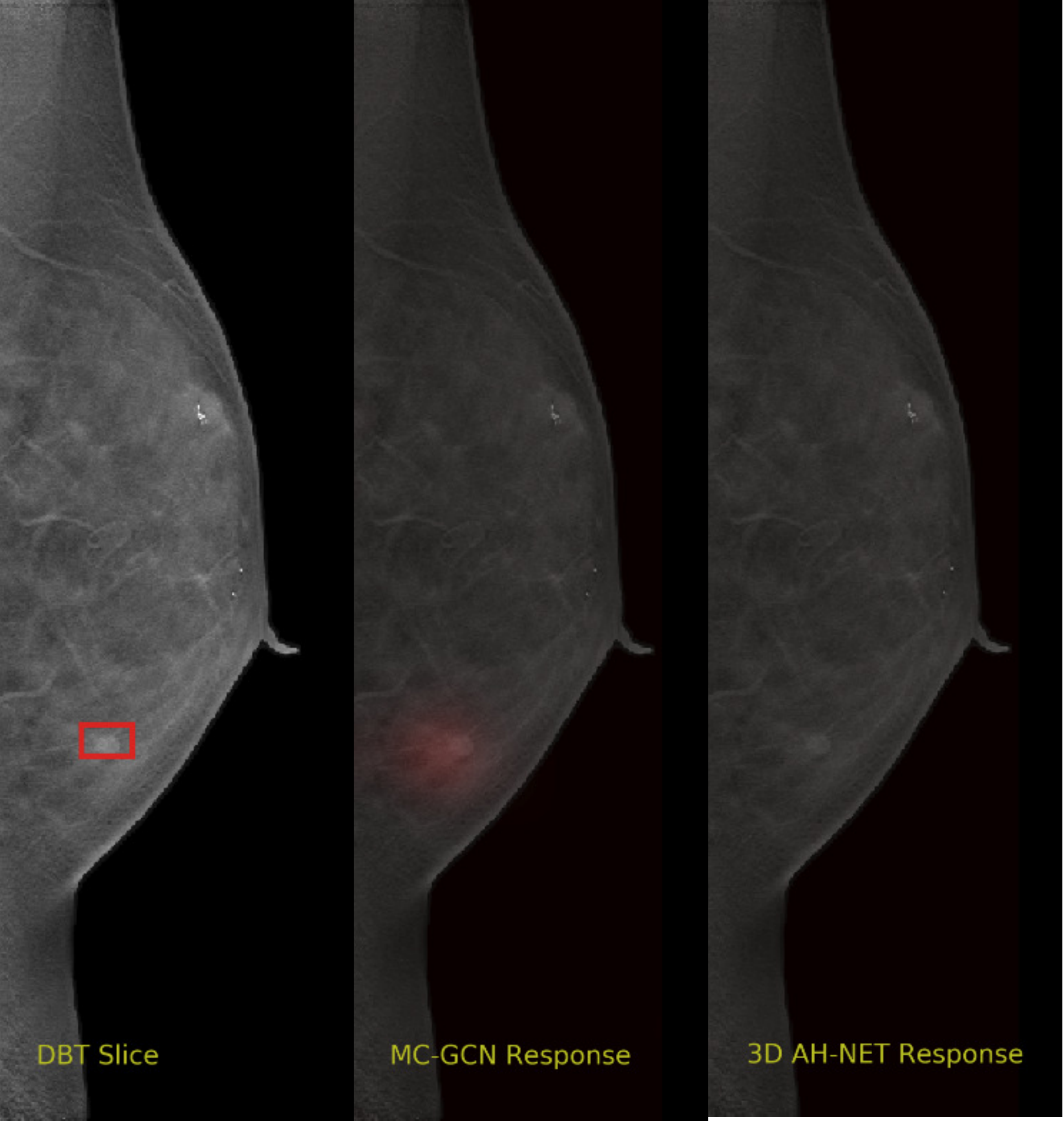}
\caption{Example DBT slice 15 with a lesion that neither network is able to detect. It is hard for the network to distinguish the lesion from the lymph nodes or skin moles.}
\label{fig:DBT-C5}
\end{figure}

We selected some example slices from the DBT dataset to demonstrate the advantage of our proposed AH-Net for the Breast cancer screening.
From Fig.~\ref{fig:DBT-A1} to Fig.~\ref{fig:DBT-A5}, we show slices from five test DBT volumes that both the MC-GCN and the proposed 3D AH-Net could successfully detect the suspected breast lesion. The original DBT slice is shown on the left with the lesion annotated by our radiologist. Please note the original annotation is a 3D box.
The figures in the middle and on the right are response maps from MC-GCN and 3D AH-Net  overlaid on the original image, respectively. The detection locations obtained with non-maximal suppression are displayed with cross markers.
As shown in the images, the proposed network can detect breast lesions varying in sizes and appearances. The confidence of the 3D AH-Net is usually higher than that of MC-GCN.
From Fig.~\ref{fig:DBT-B1} to Fig.~\ref{fig:DBT-B5}, we show five volumes that MC-GCN failed to detect the lesions since the lesions were not distinguishable from other breast tissues using the information within the slice. In contrast, 3D AH-Net was able to detect the lesions from such volumes using the 3D context between slices.
As shown in Fig.~\ref{fig:DBT-C1} to Fig.~\ref{fig:DBT-C5}, there are also volumes with lesions that both network failed to detect. Such lesions normally reside in the dense breast tissues. The boundary between these lesions and the normal breast tissues usually have low contrast. The networks sometimes also confuse them with other roundish structures in the breast such as lymph nodes or skin moles.

\section{Liver Tumor Segmentation Challenge}
\begin{figure*}[!htb]
\includegraphics[width=1\linewidth]{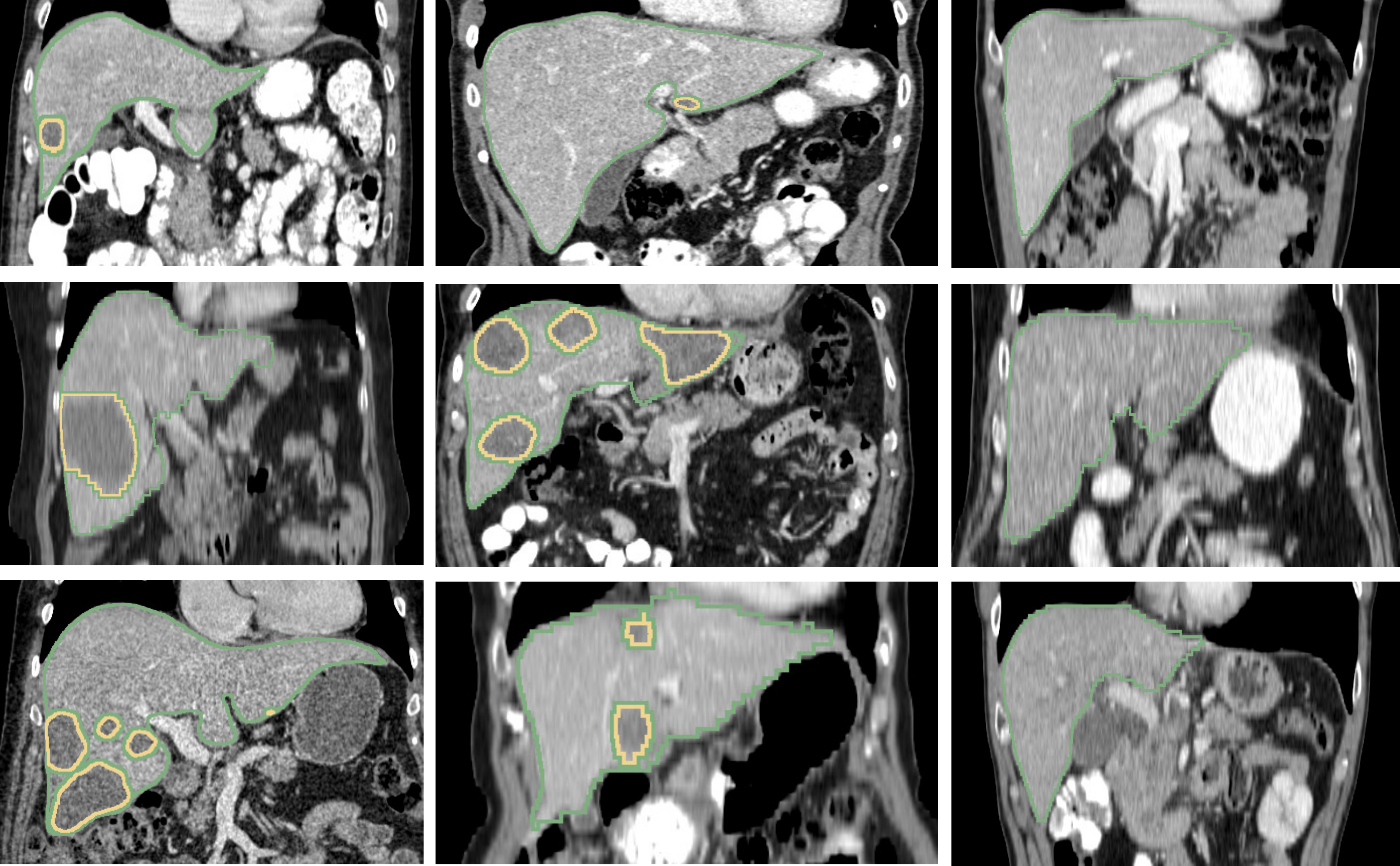}
\caption{Example sagittal view slices from the LITS challenge test volumes overlaided with the segmentation boundaries obtained with 3D AH-Net. The livers and the lesions both vary in sizes, morphology and intensities.}
\label{fig:livers}
\end{figure*}

We show 9 example sagittal slices from the LITS challenge test set in Fig.~\ref{fig:livers} to demonstrate the variation of both livers and liver lesions. The images are cropped to the region with liver centered.
The sizes and shapes of the livers vary a lot between individuals. The variation of liver lesion in sizes and intensities is even higher. The lesions are highly sparse in the abdominal CT images. Thus it is challenging for the networks to segment the lesions with small sizes. Please note that we do not have the ground truth of the test volumes. 

\begin{figure*}[!htb]
\centering
    \subfigure[Axial Image Slice]{
        \includegraphics[width=0.31\linewidth]{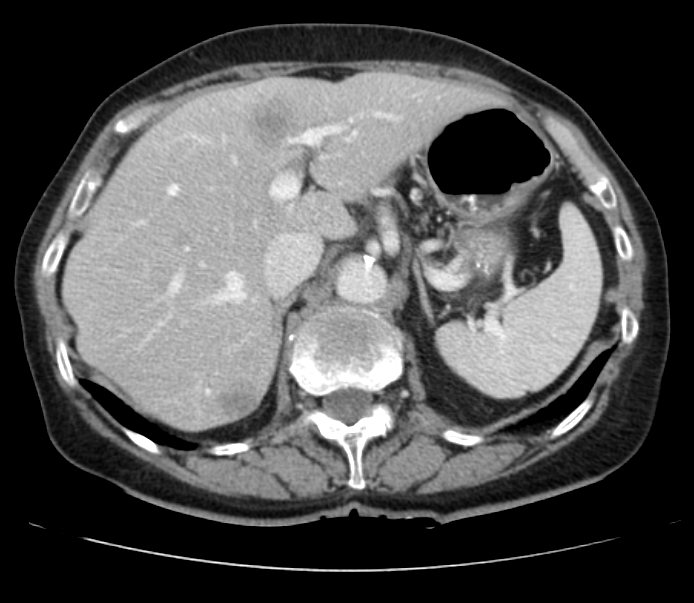}
        \label{fig:49ax-img}
    }
    \subfigure[Axial Segmentation with MC-GCN]{
        \includegraphics[width=0.31\linewidth]{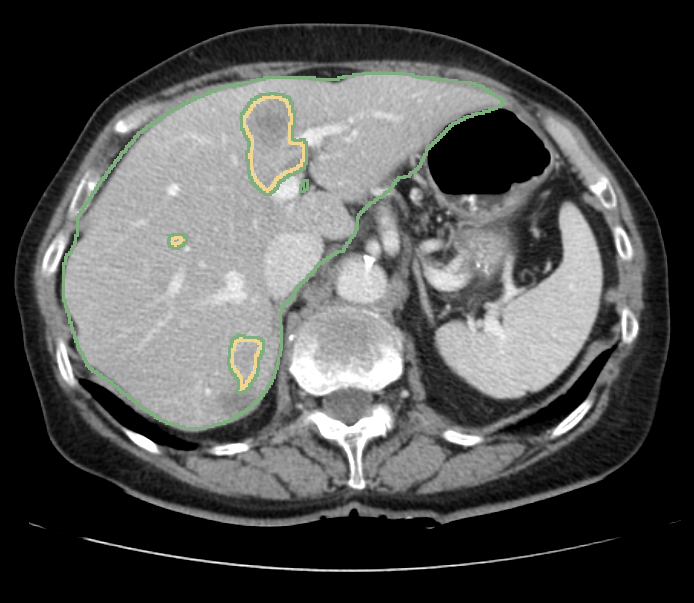}
        \label{fig:49ax-mc}
    }
    \subfigure[\bf{Axial Segmentation with 3D AH-Net}]{
        \includegraphics[width=0.31\linewidth]{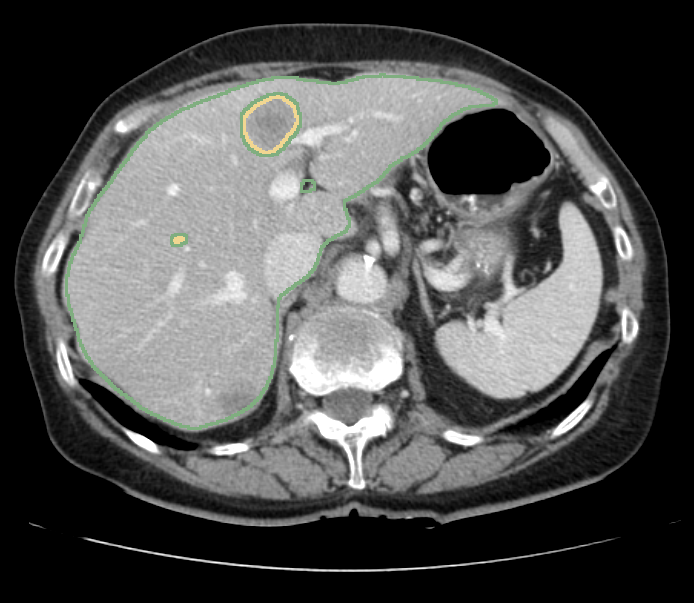}
        \label{fig:49ax-ah}
    }
    \subfigure[Sagittal Image Slice]{
        \includegraphics[width=0.31\linewidth]{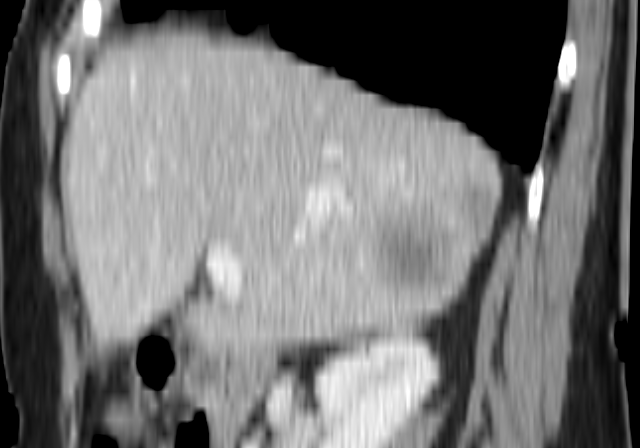}
        \label{fig:49sg-img}
    }
    \subfigure[Sagittal Segmentation with MC-GCN]{
        \includegraphics[width=0.31\linewidth]{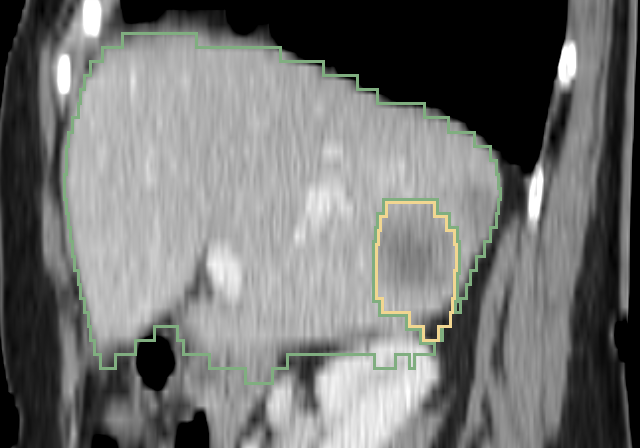}
        \label{fig:49sg-mc}
    }
    \subfigure[\bf{Sagittal Segmentation with 3D AH-Net}]{
        \includegraphics[width=0.31\linewidth]{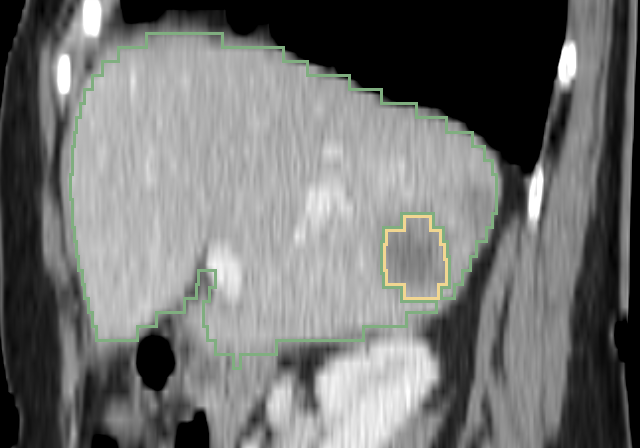}
        \label{fig:49sg-ah}
    }
    \subfigure[Coronal Image Slice]{
        \includegraphics[width=0.31\linewidth]{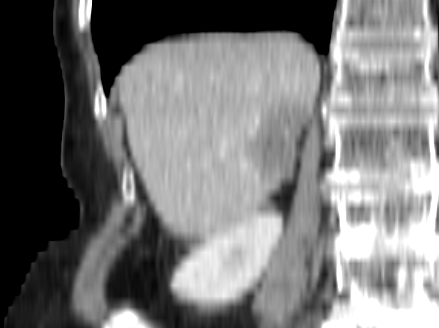}
        \label{fig:49cn-img}
    }
    \subfigure[Coronal Segmentation with MC-GCN]{
        \includegraphics[width=0.31\linewidth]{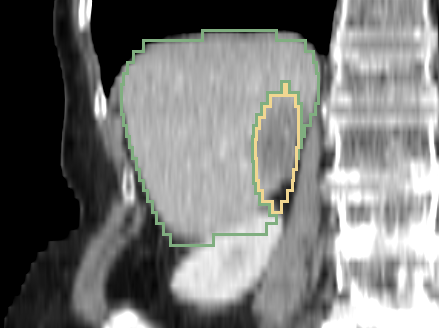}
        \label{fig:49cn-mc}
    }
    \subfigure[\bf{Coronal Segmentation with 3D AH-Net}]{
        \includegraphics[width=0.31\linewidth]{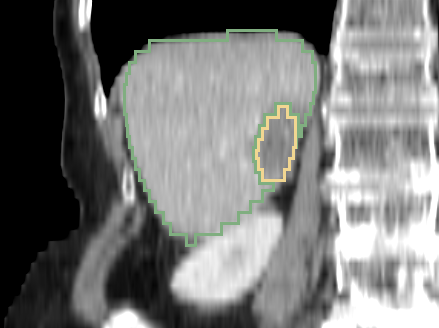}
        \label{fig:49cn-ah}
    }
    \caption{Multi-view slices from the example test CT volume 1 of the LITS challenge.}
    \label{fig:volume49}
\end{figure*}

\begin{figure*}[!htb]
\centering
    \subfigure[Axial Image Slice]{
        \includegraphics[width=0.31\linewidth]{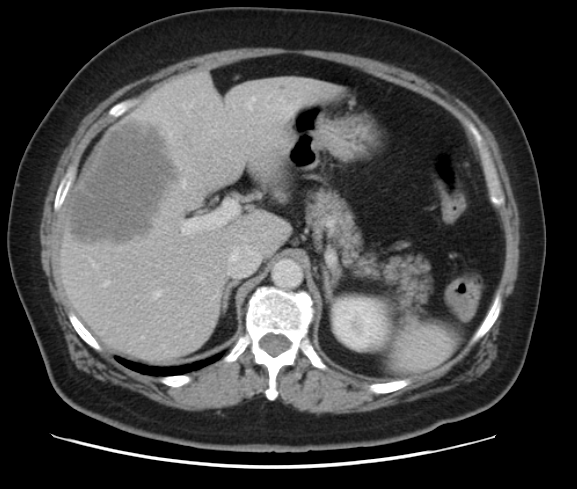}
        \label{fig:16ax-img}
    }
    \subfigure[Axial Segmentation with MC-GCN]{
        \includegraphics[width=0.31\linewidth]{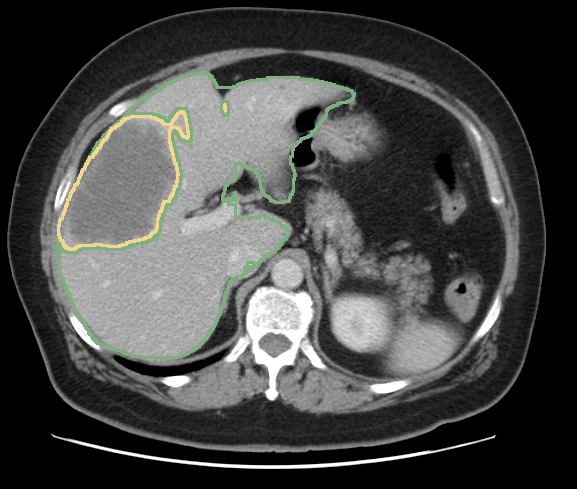}
        \label{fig:16ax-mc}
    }
    \subfigure[\bf{Axial Segmentation with 3D AH-Net}]{
        \includegraphics[width=0.31\linewidth]{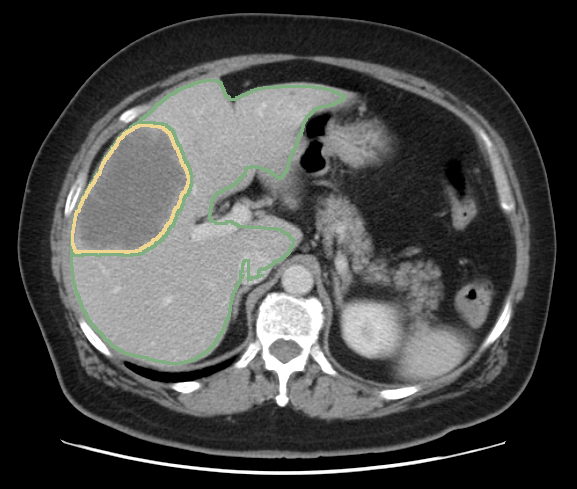}
        \label{fig:16ax-ah}
    }
    \subfigure[Sagittal Image Slice]{
        \includegraphics[width=0.31\linewidth]{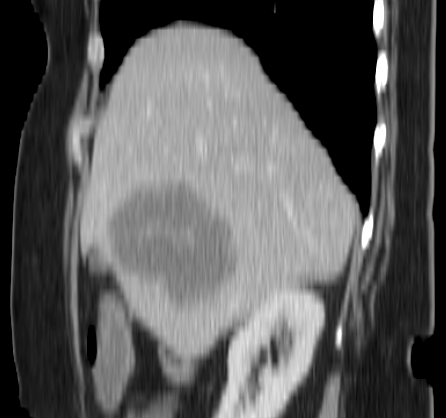}
        \label{fig:16sg-img}
    }
    \subfigure[Sagittal Segmentation with MC-GCN]{
        \includegraphics[width=0.31\linewidth]{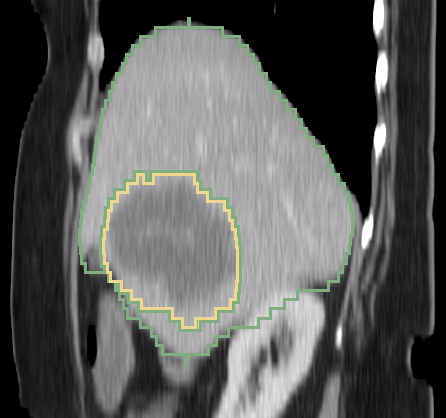}
        \label{fig:16sg-mc}
    }
    \subfigure[\bf{Sagittal Segmentation with 3D AH-Net}]{
        \includegraphics[width=0.31\linewidth]{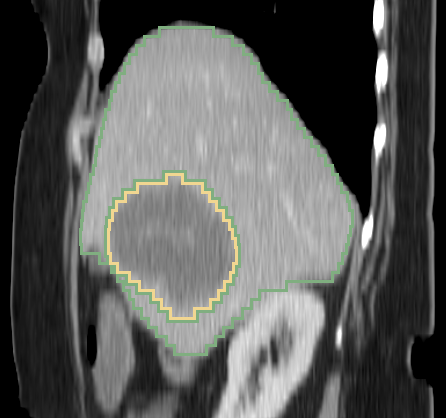}
        \label{fig:16sg-ah}
    }
    \subfigure[Coronal Image Slice]{
        \includegraphics[width=0.31\linewidth]{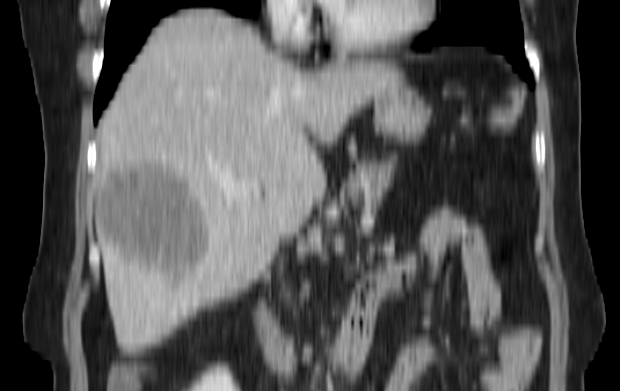}
        \label{fig:16cn-img}
    }
    \subfigure[Coronal Segmentation with MC-GCN]{
        \includegraphics[width=0.31\linewidth]{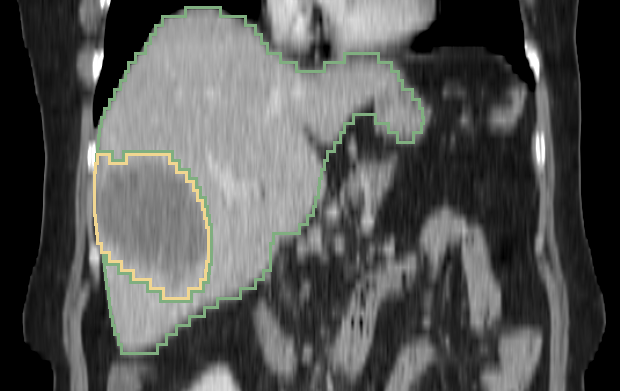}
        \label{fig:16cn-mc}
    }
    \subfigure[\bf{Coronal Segmentation with 3D AH-Net}]{
        \includegraphics[width=0.31\linewidth]{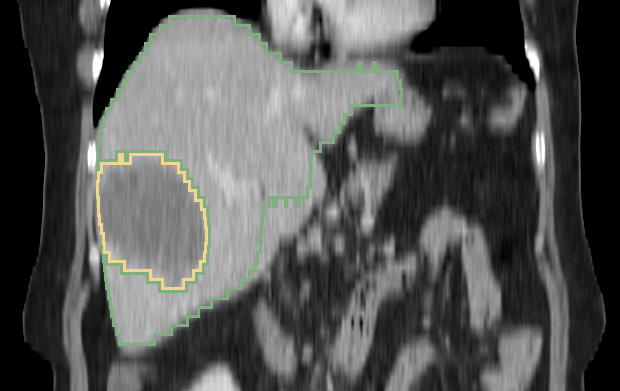}
        \label{fig:16cn-ah}
    }
    \caption{Multi-view slices from the example test CT volume 2 of the LITS challenge.}
    \label{fig:volume16}
\end{figure*}

\begin{figure*}[!htb]
\centering
    \subfigure[Axial Image Slice]{
        \includegraphics[width=0.31\linewidth]{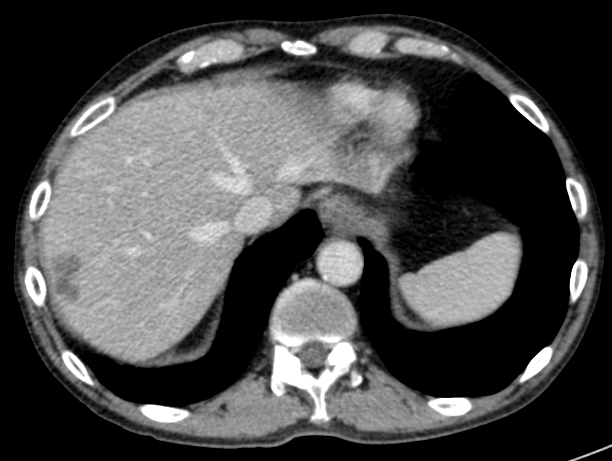}
        \label{fig:20ax-img}
    }
    \subfigure[Axial Segmentation with MC-GCN]{
        \includegraphics[width=0.31\linewidth]{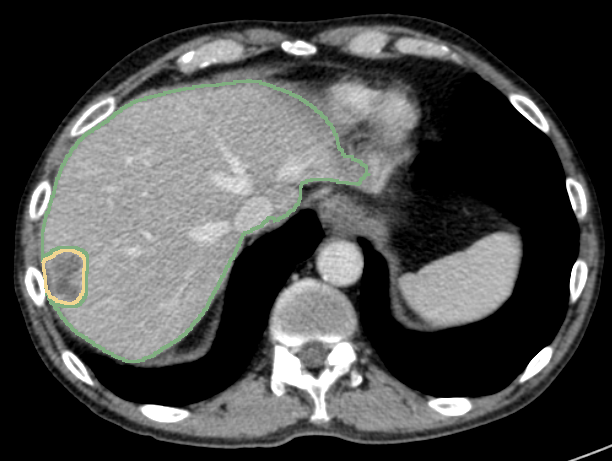}
        \label{fig:20ax-mc}
    }
    \subfigure[\bf{Axial Segmentation with 3D AH-Net}]{
        \includegraphics[width=0.31\linewidth]{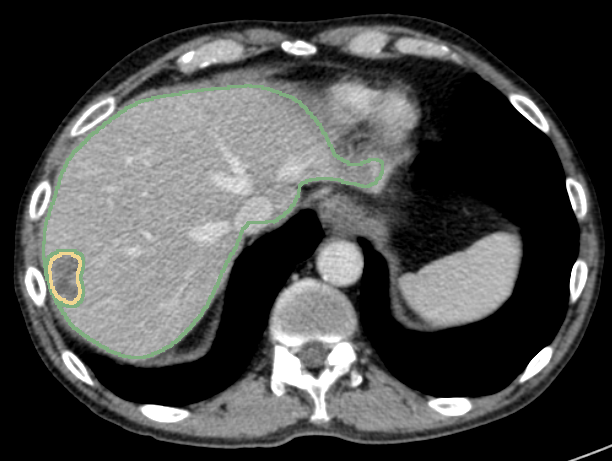}
        \label{fig:20ax-ah}
    }
    \subfigure[Sagittal Image Slice]{
        \includegraphics[width=0.31\linewidth]{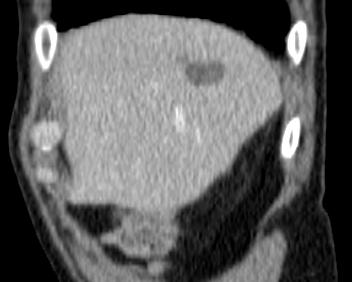}
        \label{fig:20sg-img}
    }
    \subfigure[Sagittal Segmentation with MC-GCN]{
        \includegraphics[width=0.31\linewidth]{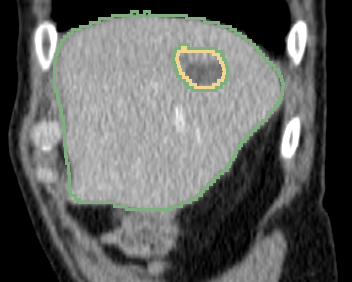}
        \label{fig:20sg-mc}
    }
    \subfigure[\bf{Sagittal Segmentation with 3D AH-Net}]{
        \includegraphics[width=0.31\linewidth]{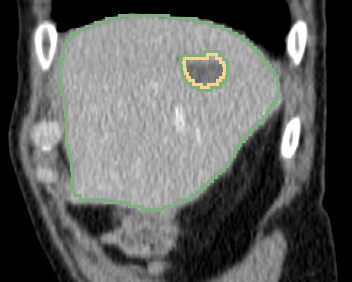}
        \label{fig:20sg-ah}
    }
    \subfigure[Coronal Image Slice]{
        \includegraphics[width=0.31\linewidth]{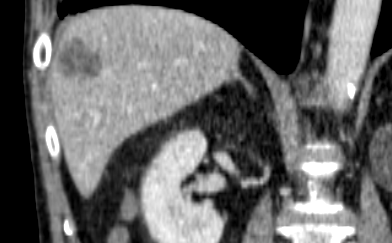}
        \label{fig:20cn-img}
    }
    \subfigure[Coronal Segmentation with MC-GCN]{
        \includegraphics[width=0.31\linewidth]{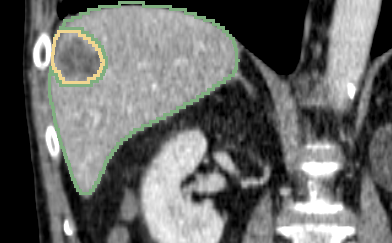}
        \label{fig:20cn-mc}
    }
    \subfigure[\bf{Coronal Segmentation with 3D AH-Net}]{
        \includegraphics[width=0.31\linewidth]{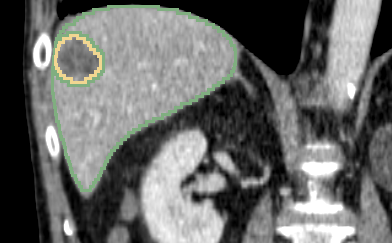}
        \label{fig:20cn-ah}
    }
    \caption{Multi-view slices from the example test CT volume 3 of the LITS challenge.}
    \label{fig:volume20}
\end{figure*}

Three example volumes are selected from the test image set to demonstrate the effectiveness of our proposed network in Fig.~\ref{fig:volume49}, Fig.~\ref{fig:volume16} and Fig.~\ref{fig:volume20}. Although we do not have the groundtruth label maps for the testing images, the liver boundaries and the presence of lesions can be visually inspected. The liver lesions normally appear as a dark region within the liver. Without sufficient 3D context, MC-GCN tends to generate false positive regions at the structure boundaries, especially under low image contrast. From the sagittal and coronal views, it is visible that MC-GCN could not generate the correct boundaries close to the top or the bottom of the lesion. By considering the consistency between slices, 3D AH-Net can segment the structures in 3D correctly, although the feature extraction network is transferred from a 2D network. The jagged boundary in the sagittal and coronal view is due to the low resolution in the z direction.

\end{document}